\documentclass[letterpaper, 10 pt, conference]{IEEEconf}  

\makeatletter

\def\@begintheorem#1#2{%
  \@IEEEtmpitemindent\itemindent
  \topsep 8pt plus 2pt minus 2pt%
  \itshape\trivlist
  \item[\hskip \labelsep{\indent\bfseries\upshape #1\ #2:}]%
  \itemindent\@IEEEtmpitemindent
}

\def\@opargbegintheorem#1#2#3{%
  \@IEEEtmpitemindent\itemindent
  \topsep 8pt plus 2pt minus 2pt%
  \itshape\trivlist
  \item[\hskip \labelsep{\indent\bfseries\upshape #1\ #2\ (#3):}]%
  \itemindent\@IEEEtmpitemindent
}

\def\@endtheorem{%
  \endtrivlist\unskip
  \vspace{2pt}%
}

\def\url#1{\href{#1}{[link]}}
\def\Url#1{\href{#1}{[link]}}

\makeatother

\IEEEoverridecommandlockouts                              
\title{\LARGE \bf
Online Design of Dynamic Networks
}

\author{Duo Wang, Andrea Araldo, Mounim El-Yacoubi
\thanks{D. Wang is at Peking University. A. Araldo and M. El-Yacoubi are at SAMOVAR, Télécom SudParis, Institut Polytechnique de Paris, 91120 Palaiseau, France.
        {\tt\small araldo@telecom-sudparis.eu}}%
}

\let\labelindent\relax
\usepackage{enumitem}

\newcommand{\keepcomment}{0} 
 \usepackage[normalem]{ulem}
 \usepackage{xcolor}
\ifnum\keepcomment=1
	\usepackage[colorinlistoftodos,textsize=scriptsize]{todonotes} 
    \newcommand{\stkout}[1]{\ifmmode\text{\sout{\ensuremath{#1}}}\else\sout{#1}\fi}
    
\else
	
	\usepackage[disable]{todonotes} 
\fi

\usepackage{tabularx}
\usepackage{subcaption}
\usepackage{amsmath}
\DeclareMathOperator*{\argsup}{argsup}
\usepackage{calrsfs} 
\DeclareMathAlphabet{\pazocal}{OMS}{zplm}{m}{n}
\usepackage{amssymb}   
\usepackage{bm}        
\usepackage{algorithm}
\usepackage{algorithmic}
\usepackage{bbm} 
\usepackage{hyperref}

\newtheorem{theorem}{Theorem}
\newtheorem{proposition}[theorem]{Proposition}
\newtheorem{lemma}[theorem]{Lemma}

\newtheorem{assumption}[theorem]{Assumption}
\newtheorem{remark}[theorem]{Remark}
\newtheorem{corollary}[theorem]{Corollary}
\usepackage{multirow}

\usepackage{layout}

\usepackage{dblfloatfix}

\begin{document}

\bstctlcite{IEEEexample:BSTcontrol} 

\maketitle
\thispagestyle{empty}
\ifnum\keepcomment=1
    \pagestyle{plain}
\else
    \pagestyle{empty}
\fi

\begin{abstract}
  Designing a network, such as a telecommunication or transport network, is typically performed \emph{offline}, prior to operation, which limits its adaptivity to stochastic variations of the environment. This paper introduces the first method to design dynamic networks \emph{online}, to adapt in real time to the environment to preserve performance.

  We formulate an impulsive control strategy and, under natural assumptions, we optimize it via a Markov-Decision Process-based online control, which designs the network on-the-fly, during its operation. Two major challenges emerge: (i)~The state-action space exhibits combinatorial explosion, due to the many sequential construction trajectories made possible over time, (ii)~The need to schedule parts of the network in advance, to enable planning tasks, e.g., shortest path queries. To tackle~(i), we guide the exploration of a Monte Carlo Tree Search toward interesting network trajectories via a neural network policy, trained via imitation learning on a limited set of exemplary dynamic networks. To tackle~(ii), we resort to a prediction model simulating future environment dynamics. As a use case, we show that our approach enables the operation of a futuristic dynamic public transport network, where bus lines are constructed on the fly in response to stochastic user demand. 
  Numerical results on a realistic scenario show efficiency superior to state-of-the-art vehicle routing problem methods, which incrementally extend individual vehicle trajectories, lacking network-wide structure. Additional case studies on complex system management and the $k$-server problem demonstrate the generality of our method.
\end{abstract}

\section{Introduction}
\label{section_Introduction}
\todo{update the arxiv link}
\todo{aa: add sketches of the proof}
Networks often exhibit temporal dynamics: social ties evolve over days, public transport lines vary across time of day, etc. These dynamics are captured by Temporal Graphs~(TGs). 
\todo{The refs in the introduction and related work are richer in the ICAPS26 submission}
Prior research has primarily focused on \emph{observing} changes in TGs, to summarize or predict the underlying patterns~\cite{fournier2020survey}. In contrast, we here focus on \emph{designing} the dynamics of TGs, which is relevant when we can control a network, e.g., of computing nodes, of bus lines, or of potential process threads. In contrast with previous descriptive approaches~\cite{fournier2020survey}, our focus is prescriptive. The goal is to decide the evolution of TGs to maintain some performance target, faced with an environment that generates random requests, to be routed over the designed network, or random costs on its edges. 
%
Our main contributions are:

\begin{enumerate}[leftmargin=*, labelsep=0.3em]
\item To our knowledge, this paper presents the first systematic formalization of the problem of \emph{designing} dynamic networks \emph{online}, during their operation, in a stochastic and unknown environment. This is in contrast to previous work, in which (i)~networks are designed offline, i.e., planned prior to their operations, or in which (ii)~dynamic networks are observed and analyzed, but are not subject to decision-making.

\item We find the conditions under which the problem can be solved via a discrete time Markov Decision Processes.

\item We provide a practical solution method, based on Monte-Carlo Tree Search~(MCTS). MCTS typically requires extensive exploration to learn an effective Q-function~\cite{hamrickcombining}, which is unsuited for our online setting, in which decisions must be taken on-the-fly. To address this, we train a neural network policy via \emph{imitation learning} using a limited set of exemplary evolutions, and we use it to guide MCTS.
\end{enumerate}

\noindent After reviewing the related work~(\S{}\ref{eq:related-work}), in \S{}\ref{sec:model} network design is formalized as an impulsive control problem. We analytically prove that, to maximize the number of requests, decisions must be made at specific discrete time steps (Prop.~\ref{prop:timing}, Cor.~\ref{cor:intervention-instants}). This allows us to describe the problem in terms of Markov Decision Processes~(\S{}\ref{sec:mdp}). \S{}\ref{sec:solution-method} presents the solution method. \S{}\ref{sec:experimental-results} presents an application to the transport domain, where a stochastic sequence of incoming trip requests must be served by a fleet of buses.  In contrast to current Dynamic Vehicle Routing Problem~(DVRP) methods, where vehicle trajectories are extended one by one, lacking an overall structure, our method designs a structured \emph{network} of bus lines, enabling efficient user journeys, including transfers from one line to another. Thanks to this network-wide structure, we nearly double the requests served compared to state-of-the-art (SOTA) DVRP methods, on a dataset of real-world New York trips. Other two applications, beyond transport, showcase the generality of our method.

\section{Related Work}
\label{eq:related-work}

Designing a network is a classical offline combinatorial optimization problem. 
In more recent online settings~\cite{chekuri2025streaming}, edge propositions arrive over time and must be accepted or rejected. However, (i)~the constructed network remains static, and (ii)~the demand is assumed to be static and fully known. We instead (i)~aim to design a dynamic network, under (ii)~an unknown demand, composed of requests that arrive over time. Despite potential promising applications in several engineering domains~\cite{Schmid2025}, systematic methods to undertake such a design task are still missing.

Dynamic networks can be modeled as Temporal Graphs~(TGs). 
TGs can be represented either (1)~as a sequence of snapshots (each being a static graph) at discrete time steps or (2)~as a Time-Expanded Graph (TEG), which replicates every node at each time instant to expand the time dimension, e.g.~\cite[\textsuperscript{Fig.2B}]{FORTIN201618}.
The issue with~(1) is does not capture changes between the steps. We here focus on TEGs~(2), since they can capture changes occurring at any moment and are used to model practical systems~\cite{FORTIN201618}.

Recent works typically assume that the TEG is given as an exogenous input and solve optimization problems over it, e.g., how many resources to deploy over the TEG~\cite{Hewitt2020}, how to compute shortest paths~\cite{Helmberg2014,Raviv2022}, or connectivity structures~\cite{deligkas2025many}. Generally, the demand is assumed to be fully known.

Two main aspects make our work fundamentally different: (i)~The TEG is not an input for us; it is rather a control variable, and we have to decide which edges to add and when to add them. (ii) Methods~\cite{Hewitt2020,Raviv2022} are \emph{offline}, i.e., all computations are done in a planning phase, prior to operating the network. Our method is instead \emph{online}, i.e., it decides the TEG representing the network while the network is running.

\section{Model and Problem Formulation}
\label{sec:model}
\noindent We design overlay dynamic graph~$\pazocal G(t)$ (\S{}\ref{sec:Time-expanded-graph}) over an input static substrate graph~$\pazocal G_\text{substr}$. A random environment generates requests that must be routed on~$\pazocal G(t)$~(\S{}\ref{sec:environment-and-routing}). The environment determines some transitions of the system state~(\S~\ref{sec:state-transitions}), which we can control via an impulsive strategy~\ref{sec:control}. To simplify the optimization of such strategy, we prove that the system can be described in terms of a time-stepped Markov Decision Process~(\S{}\ref{sec:mdp}). Due to the page limit, proofs are in our extended report~\cite[\textsuperscript{App.\ref{sec:proofs}}]{WangExtended}.

\subsection{Substrate and overlay time-expanded graphs}
\label{sec:Time-expanded-graph}
We formulate the problem of designing a dynamic network as a sequence of decisions to construct a Time Expanded Graph~(TEG), lying onto a \emph{substrate graph}
$$\pazocal{G}_\text{substr}=(\pazocal{V},\pazocal{E}_\text{substr},\{w_{u,u'}\}_{(u,u')\in\pazocal E_\text{substr}},\{c_{u,u'}\}_{(u,u')\in\pazocal E_\text{substr}}),$$where~$\pazocal{V}$ is a set of nodes and~$\pazocal{E}_\text{substr}\subseteq\pazocal{V}\times\pazocal{V}$ the set of edges.
If edge~$(v,v')\in\pazocal{E}_\text{substr}$, then it is possible to go directly from node~$v$ to node~$v'$, taking a certain deterministic time~$w_{u,u'}\ge 0$ and incurring a certain stochastic cost~$c_{u,u'}\ge 0$. 

Substrate graph~$\pazocal G_\text{substr}$ is not subject to our optimization; it is instead an exogenous input. We want to build a dynamic network~$\pazocal G(t),t\in[0,T]$ on top of~$\pazocal G_\text{substr}$. For instance, $\pazocal G_\text{substr}$ may represent the road network and~$\pazocal G(t)$ some bus routes built on top of it.
A table of notation is in our extended report~\cite[\textsuperscript{Table~\ref{tab:notation}}]{WangExtended}.


A Time-Expanded Graph (TEG) is a pair~$\pazocal G(t)=(\pazocal G_\text{substr},\pazocal E(t))$, where~$\pazocal E(t)$ is a set of \emph{time-expanded edges}. 
A time-expanded edge is a tuple~$e=(\tau,v,v')$ (with $(v,v')\in\pazocal E_\text{substr},\tau\in]0,T[$). It indicates that the designed TEG allows transitioning from node~$v$ to~$v'$, leaving at instant~$\tau$ and arriving at time~$\tau'=\tau+w_{v,v'}$, which we call \emph{departure time} and \emph{arrival time} of~$e$, respectively. Arrival time and cost are determined exogenously by substrate graph attributes~$w_{v,v^\prime}, c_{v,v^\prime}$.\footnote{
We simply write ``edge'' when the context or notation makes clear whether it refers to a substrate edge or a time-expanded edge.
}
Set~$\pazocal E(t)$ contains all the time-expanded edges that have been added up to time~$t$ and that are still active, i.e., their departure time is no earlier than~$t$. Time-expanded edges are added one by one. If edge~$e=(\tau,v,v')$ is added in the TEG at instant~$t$, i.e., $\{e\}=\pazocal E(t^+)\setminus\pazocal E(t^-)$,\footnote{
As usual, $\pazocal E(t^-),\pazocal E(t^+)$ denote the left and right limits, i.e., the set of time-expanded edges before and after the insertion of the new edge.
} it can be used to route requests, only during interval $]t,\tau[$. 

For greater generality, we consider the case in which~$\pazocal G(t)$ is a multilayer time-expanded graph, i.e., it is composed of a set~$\pazocal L$ of layers, each of which is a time-expanded graph: 
\begin{align}
\label{eq:multiple-layers}
\pazocal G(t)=\left( \pazocal G_\text{substr}, \pazocal E(t) \right); 
&&
\pazocal E(t) = \bigcup\nolimits_{l\in\pazocal L} \pazocal E_l(t)
\end{align}

We call time-expanded graph~$\pazocal G_l(t)=\left( \pazocal G_\text{substr}, \pazocal E_l(t) \right)$ a \emph{layer}. We can thus model complex networks, composed of multiple components working simultaneously, each component represented by a layer. Examples are a public transport network resulting from the movement of multiple vehicles, or a multi-server system, where each server can execute a separate sequence of tasks, or an organization where multiple potential threads of activities can occur. 

Ass.~\ref{ass:sequence} implies that each system component executes activities in sequence and that we continue designing a network until all layers cover the entire lifespan~$T>0$.

\begin{assumption}
\label{ass:sequence}
Each layer~$\pazocal G_l(t),l\in\pazocal L$ is a consecutive sequence of time-expanded edges, i.e., the arrival time and node of one edge in~$\pazocal G_l(t)$ corresponds to the departure time and node of the next. The design terminates when
\begin{align}
\label{eq:lifespan}
\forall l\in\pazocal L \,\,\, \exists e=(\tau,v,v')\in\pazocal \pazocal G_l(t): \tau+w_{v,v'}>T.
\end{align}
\end{assumption}

\subsection{Environment and routing}
\label{sec:environment-and-routing}

Environment~$\textit{Env}$ is a marked point process over sample space~$\Omega$~\cite[\textsuperscript{Def.~2.1.2}]{Jacobsen2006}. For any realization~$\omega\in\Omega$, $\textit{Env}$ generates a sequence~$\pazocal R_\omega$ of requests. $\pazocal R_\omega$ is exogenous, i.e., we have no influence on it. A request~$d=(\tau,v,v')\in\pazocal R_\omega$ denotes a request arrived at instant~$\tau\in]0,T[$ for a service departing at node~$v\in\pazocal V$ and ending at node~$v'\in\pazocal V$.
Request~$d=(\tau,v,v')$ is routed (or, equivalently, served) on TEG~$\pazocal G(t)$ if there exists a time-expanded \emph{path}~$p=(e_1,\dots,e_m)$, where~$e_1,\dots,e_m$ is a set of time-expanded edges in~$\pazocal G(t)$, which satisfies:
\begin{align}
    \label{eq:path-constraint-start}
    \tau
    &\le \tau_1
    & 
    \\
    e_j
    &= (\tau_j, v_j, v_{j+1})\in\pazocal{E}(t),
    \,\,\,\ 
    \forall j=1,..,m
    \\
    \label{eq:path-continuity}
    \tau_j + w_{v_j, v_{j+1}}
    &
    \begin{cases}
    = \tau_{j+1} & \text{ if }e_j,e_{j+1}\text{ in the same layer }l\\
    \le \tau_{j+1} & \text{ otherwise}
    \end{cases}
    \nonumber
    \\
    &
    \,\,\,\,\,\,\,\,\,\,\,\,\,\,\,\,\,\,\,\,\,\,\,\,\,\,\,\,\,\,\,\,\,\,\,\,\,\,\,\,\,\,\,\,\,\,\,\,\forall j=1,\dots,n-1
    \\
    v_1=v, & \,\,\,\ v_{m+1}=v'
    \label{eq:final-node}
\end{align}
Constraints~\eqref{eq:path-constraint-start}-\eqref{eq:final-node} ensure the geometrical continuity of path~$p$. Note that~\eqref{eq:path-continuity} allows a path to go through multiple layers, in which case it suffices that the arrival time of the last edge traversed in a layer is no later than the departure time of the first edge traversed in the other layer. To ensure continuity, the sections of a path contained into a single layer must have all edges connected in time (the arrival time of one must correspond to the departure time of the other).
\begin{align}
\label{eq:arrival-time}
AT(p) &:=\tau_n + w_{v_n, v'} \text{ (arrival time of path $p$)}
\\
AT(p)-t & \le L \text{ (maximum admissible path length)}
\label{eq:max-travel}
\end{align}
Eq.~\eqref{eq:arrival-time} defines the arrival time of path~$p$, and~\eqref{eq:max-travel} imposes that a path must not be longer than time~$L$ to be considered feasible. Let~$\pazocal P(t,v,v')$ denote the set of feasible paths available at time~$t$, i.e., paths within time-expanded graph~$\pazocal G(t)$ and respecting~\eqref{eq:path-constraint-start}-\eqref{eq:max-travel}. All the analysis is solely based on whether, for a request~$d=(\tau,v,v')$, at least one feasible path exists, no matter the criterion to select the path, if multiple exist. For simplicity, in the results, we choose the shortest path, but any other criterion would work.

\subsection{State transitions}
\label{sec:state-transitions}
Let~$d=(\tau,v,v')$ be a request arrived at time~$\tau$, requesting a service from node~$v$ to node~$v'$. There can be two cases:\\
(a)~There is already some feasible path in the graph, i.e., $\pazocal P(\tau,v,v')\neq \emptyset$. In this case, the request is immediately served.\\
(b) Otherwise, the request is added to a buffer~$\pazocal B(t)$, with~$t=\tau$, i.e., $\pazocal B(\tau^+)=\pazocal B(\tau^-)\cup \{d\}$.

We call pair~$s(t)=\left(\pazocal B(t),\pazocal G(t)\right)$ the \emph{state of the system}. The decision policy consists of adding time-expanded edges over time into~$\pazocal G(t)$, and will be discussed in \S{}\ref{sec:control}. Let us first describe the state transitions without any control decisions: the state changes at instant~$t\in]0,T[$ if one of the following events occurs:\\
(i)~A new request~$d=(\tau,v,v')$ arrives at instant $\tau=t$ and is inserted into the buffer (see case~(b) above). 
\\
(ii)~One of the edges expires in~$t$, i.e, $\exists e=(\tau,v,v')\in\pazocal E(t^-):\tau=t$; in this case~$\pazocal E(t^+)=\pazocal E(t^-)\setminus\{e\}$.

Observe that both events change state~$s(t)$: in particular, event~(i) changes~$\pazocal B(t)$, event~(ii) changes~$\pazocal G(t)$.

\begin{proposition}
\label{prop:PDMP}
In the absence of any control action, state~$s(t)$ evolves as a Piecewise Deterministic Markov Process~(PDMP) \cite[\textsuperscript{Def.1}]{Sabbadin2025} \end{proposition}

\subsection{Control}
\label{sec:control}
Prop.~\ref{prop:PDMP} provides the basis for employing \emph{Impulsive Control Strategies}~(ICS), which are well studied for PDMPs \cite[\textsuperscript{Def.4}]{Sabbadin2025}. An ICS is denoted by~$\textit{ICS}=(t_n,e_n)_{n\ge 1}$: at instants~$t_n$, called \emph{intervention times}, we add time-expanded edge~$e_n=(\tau_n,v_n,v_n')$ to~$\pazocal G(t)$, i.e.,\footnote{
From the PDMP formulation of~\cite{Sabbadin2025} derives that the probability that two events (PDMP jumps (\S{}\ref{sec:state-transitions}) or interventions) occur at the exact same instant is null.
} 
\begin{align}
\label{eq:change-of-E}
\pazocal E(t_n^+)=\pazocal E(t_n^-)\cup\{e_n\}, \forall n\ge 0.
\end{align}
An ICS decides \emph{which} time-expanded edges to add and \emph{when}. In the presence of multiple layers~\eqref{eq:multiple-layers}, time-expanded edges are added into one among~$\pazocal E_l(t),l\in\pazocal L$.

Inserting edge~$e_n$ at time~$t_n$ can enable new paths, i.e., $\pazocal P(t_n^+,v,v)\supseteq \pazocal P(t_n^-,v,v)$.
Therefore, some requests waiting in buffer~$\pazocal B(t_n^-)$ may be served after adding~$e_n$. Denoting such requests by~$\pazocal B'=\{d=(\tau,v,v')\in\pazocal B(t_n^-)|\pazocal P(t_n^+, v, v')\neq \emptyset\}$, we have
\begin{align}
\label{eq:change-of-B}
\pazocal B(t_n^+) = \pazocal B(t_n^-)\setminus \pazocal B'.
\end{align}

The action consisting of adding edge~$e_n$ at instant~$t_n$ thus changes the state via~\eqref{eq:change-of-E} and~\eqref{eq:change-of-B}.

Let~$\pazocal A(t_n,s(t_n))$ denote the set of candidate time-expanded edges, i.e., those that are feasible to add. Such a set might be restricted to represent some real-world constraints. To meet Ass.~\ref{ass:sequence}, we here impose that time-expanded edge~$e_n$ can be added to a layer~$l\in\pazocal L$ only if it is connected to the last preexisting edge in~$\pazocal G_l(t_n)$.

We consider two indicators to evaluate a strategy. For any realization~$\omega\in\Omega$ a realization, the served requests during lifespan~$[0,T]$ are those for which a path is find at some instant~$t$. The \emph{number of served requests} is thus
\begin{align}
\label{eq:num_served_req}
N_\omega^\text{ICS}
:=
&\left|
\left\{
\begin{array}{l}
d=(\tau,v,v')\in\pazocal R_\omega 
\Large|
\\
\exists t\in]\tau,T[: \pazocal P(v,v',t)\neq \emptyset
\end{array}
\right\}
\right|,
\end{align}
where~$|\cdot|$ is the cardinality of the set. Observe that, by modifying~$\pazocal E(t)$ (and thus~$\pazocal G(t)$), strategy $\textit{ICS}$ also modifies~$\pazocal P(v,v',t)$, which thus impacts~$N_\omega^\text{ICS}$.

The \emph{cumulative cost} is
\begin{align}
\label{eq:cumulative-cost}
C_\omega^\textit{ICS}
:=
\sum_{n\ge 1} c_{v_n,v_n',\omega}
\end{align}
where~$c_{v_n,v_n',\omega}$ is the realized value of cost~$w_{v_n,v_n'}$ of the substrate edge~$(v_n,v_n')\in\pazocal E_\text{substr}$ used by time-expanded edge~$e_n=(\tau_n,v_n,v_n')$.

Compared to classical impulsive control strategies, here we take actions (consisting of adding time-expanded edges) that have a time-related feature (which is the departure time of such time-expanded edges). This helps reduce the space of ``good'' decisions, whose timing is determined by maximum path length~$L$ (defined in~\eqref{eq:max-travel}), via the following proposition.

\begin{proposition}
\label{prop:timing}
The maximum number of served requests is achieved by a strategy~$\textit{ICS}=\left(t_n, e_n=(\tau_n,v_n,v_n')\right)_{n\ge 0}$ such that
$t_n=\tau_n-L$.
\footnote{
The full proof is in our extended report~\cite[\textsuperscript{App.~\ref{sec:proofs}}]{WangExtended}. We provide here a sketch.
Consider any admissible strategy $(t_n, e_n=(\tau_n, v_n, v'_n))_n$ and a modified one where the insertion time $t_n$ is shifted earlier to $t'_n < t_n$, keeping the same departure time $\tau_n$.
For any $(v,v')$ and time $t$, the path set of this latter strategy is a superset of the path set of the original admissible strategy. This is because inserting an edge earlier enlarges the set of feasible time-expanded paths, by making them available earlier.
Therefore, the number of served requests cannot decrease when anticipating decisions.
However, paths cannot be too long (constraint~\eqref{eq:max-travel}) and it is thus sufficient to schedule edges exactly $L$ time units in advance, without loss of optimality.
}
\end{proposition}

\begin{remark}
\label{rem:restriction}
In this paper, we thus restrict our attention to the strategies complying with Prop.~\ref{prop:timing}. 
\end{remark}

\begin{corollary}
\label{cor:intervention-instants}
Suppose that a strategy complies with Prop.~\ref{prop:timing} and Ass.~\ref{ass:sequence}. Then, there exists a deterministic function~$f^\text{next}(\cdot)$ that, given intervention instant~$t_n$ and the time-expanded graph~$\pazocal G(t_n)$ at that instant, returns the next intervention instant~$t_{n+1}$, i.e., $t_n=f^\text{next}(t_n,\pazocal G(t_n))$.
\end{corollary}

Cor.~\ref{cor:intervention-instants} removes the burden of deciding \emph{when} to add edges, since such instants derive deterministically from the departure times. We can focus on just choosing \emph{which} edges must be added. This greatly simplifies the problem, whose decisions are to be taken in determined discrete time-steps.

\subsection{Markov Decision Process (MDP)}
\label{sec:mdp}
We now aim to find a policy~$\pi(\cdot)$ takes an action (at determined intervention instants), consisting of adding a time-expanded edge. Specifically $\pi(e|s(t_n))$ denotes the probability of adding time-expanded~$e\in\pazocal A(t_n, s(t_n))$, with~$\sum_{e\in\pazocal A(t_n, s(t_n))} \pi(e|s(t_n))=1$. Set~$\pazocal A(t,s(t))$ of feasible candidate edges thus corresponds to the action set.

Observe that any realization~$\pi_\omega(\cdot), \omega\in\Omega$ of policy~$\pi(\cdot)$ induces an impulsive control strategy~$ICS_{\pi_\omega}=(t_n, e_n)_{n\ge 0}$. We can thus calculate the served requests~\eqref{eq:num_served_req} and the cumulative cost~\eqref{eq:cumulative-cost} induced by a certain policy realization and their expectation induced by that policy:
\begin{align}
\label{eq:N-omega-pi}
N_\omega^\pi := N_\omega^{ICS_{\pi_\omega}} &&
C_\omega^\pi := C_\omega^{ICS_{\pi_\omega}}
\\
\label{eq:N-pi}
N^\pi := \mathbb E_\omega[N_\omega^\pi] &&
C^\pi := \mathbb E_\omega[C_\omega^\pi].
\end{align}

The instantaneous reward can be defined depending on the indicator we want to optimize. Let us consider a realization~$\omega\in\Omega$ and assume that at intervention instant~$t_n$ the action chosen is a certain time-expanded edge~$e_n=(\tau_n,v_n,v_n')\sim \pi(\cdot|s(t_n))$. 

If we aim to maximize the number of served requests~\eqref{eq:num_served_req}, we define the reward as the number of requests arriving in the next time interval~$]t_n,t_n+1$[ that are directly served, plus the number of requests that were waiting in the buffer and that can be served thanks to the new addition, i.e.,
\begin{align}
\label{eq:reward-served}
&r_\omega (t_n, e_n| s(t_n))
=
\\&\nonumber
\left|
\left\{
\begin{array}{cl}
d=(\tau,v,v')\in\pazocal R_\omega |
& \tau\in]t_n,t_{n+1}[,
\\
\nonumber
&  \exists t\in]\tau,t_{n+1}]: \pazocal P(v,v',t)\neq \emptyset
\end{array}
\right\}
\right|
\\
\nonumber
& + 
\left|
\{d=(\tau,v,v')\in\pazocal B_\omega(t_n)| \pazocal P(v,v',t_n)\neq \emptyset\big\}
\right|
\end{align}

If instead we aim to minimize the cumulative cost~\eqref{eq:cumulative-cost}, we define the instantaneous reward as
\begin{align}
\label{eq:reward-cost}
r_\omega(t_n,e_n|s(t_n))=
- c_{v_n, v_n',\omega}.
\end{align}

\noindent We seek policies maximizing the expected cumulative reward
\begin{align}
\label{eq:max}
\max_\pi & \mathbb E_\omega \left[\sum_{t_n\le T} r_\omega(t_n,e_n|s(t_n))\right]
\\
\label{eq:model-constraint-instants}
\text{s.t. } &
(t_n)_{n\ge 0} \text{ are determined as in Cor.~\ref{cor:intervention-instants}}
\\&
e_n=(\tau_n, v,v')\sim \pi(\cdot | s(t_n))
\\&
t_n = \tau_n-L \text{ (Prop.~\ref{prop:timing})}
\\
\label{eq:model-constraint-state}
&
s(t_n^+) \to s(t_{n+1}^-) \text{ as described in \ref{sec:state-transitions}}.
\\
\label{eq:model-constraint-state-controlled}
&
s(t_n^-) \to s(t_n^+) \text{ as described in \ref{sec:control}}.
\end{align}

\begin{proposition}
\label{prop:objective}
Objective~\eqref{eq:max} is equivalent to maximizing the expected served requests $N^\pi$ or minimizing the expected cumulative cost~$C^\pi$, depending on whether the instantaneous reward is defined as~\eqref{eq:reward-served} or~\eqref{eq:reward-cost}.
\end{proposition}

\section{Solution Method}
\label{sec:solution-method}
Thanks to the formalization of~\S{}\ref{sec:mdp}, we can perform the online design of a dynamic network via an MDP. 
Observe that at each intervention time~$t_n$, action space~$\pazocal A(t_n,s(t_n))$, consisting of the possible time-expanded edges that can be added, might be too big for Q-learning methods. In such cases, Monte Carlo Tree Search (MCTS) is preferable, since it does not require learning Q-values for all state-action pairs. Standard MCTS enjoys asymptotic guarantees under its classical assumptions~\cite[\textsuperscript{Thm.6}]{kocsis2006bandit}; here we use it as a practical online planner in a large action space, guided by a learned prior to improve sample efficiency. MCTS performed better and required less computational resources than other methods we applied in our preliminary experiments, such as actor-critic reinforcement learning and decision transformers.

\subsection{Monte Carlo Tree Search for Dynamic Network Design}
\label{sec:MCTS}

\begin{figure}
  \centering
  \includegraphics[width=0.75\linewidth]{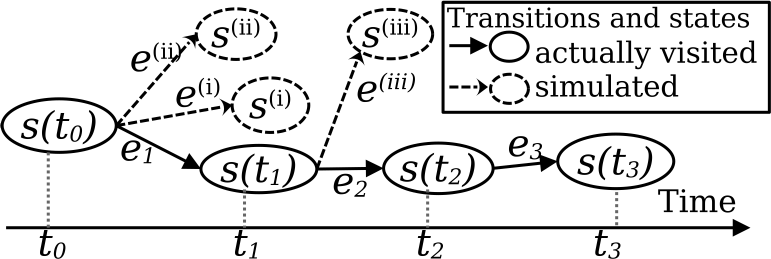}
  \caption{An example of search tree~$\textit{Tree}(s(t_0))$}
  \label{fig:Styles/NIPS_fig_1.png}
\end{figure}




MCTS constructs a search tree~$\textit{Tree}(s)$ (see Fig.~\ref{fig:Styles/NIPS_fig_1.png}), starting from current state~$s=s(t_n)=(\pazocal B(t_n), \pazocal G(t_n))$ as root.
$\textit{Tree(}s)$ is a decision tree, each node represents a new state resulting from an action.
$\textit{Tree}(s)$ is created by running~$m$ simulations. In each simulation, a sequence of random events (e.g., requests) is generated via a simulated environment~$\textit{Env}'$, which can be a prediction model like the one in \S{}\ref{sec_Requests_Generation}), mimicking the statistical characteristics of the environment. Time advancement is also simulated. 

Suppose the system is at time~$t_0$ and an action must be taken. To do so, we construct a tree via simulations. At the beginning of each simulation, we have~$\textit{Tree}(s(t_0))$, which is the result from previous simulations. 
Each simulation includes the following four consecutive steps.\\

1. \emph{Selection:} Starting from the root~$s(t_0)=(\pazocal B(t_0),\pazocal G(t_0))$, an action is chosen in~$\pazocal A(t_0, \pazocal G(t_0))$, which corresponds to selecting a child tree node. This induces the corresponding state $s(t_1)=(\pazocal B(t_1),\pazocal G(t_1))$, where~$t_1$ is the next intervention instant (Cor.~\ref{cor:intervention-instants}) and~$\pazocal G(t_1)$ is the graph obtained by applying that action on graph~$\pazocal G(t_0)$~(Fig.~\ref{fig:Styles/NIPS_fig_1.png}). The criterion underlying the choice of the action will be discussed in \S{}\ref{sec:neural-policy}. The reward of the action is estimated, based on a simulated sequence of requests arriving between~$t_0$ and~$t_1$, together with the dynamics defined by~\eqref{eq:model-constraint-instants}-\eqref{eq:model-constraint-state-controlled}.
The selection process ends when it finds an expandable node, i.e., a tree node from which there exist actions that have never been selected in previous simulations, e.g., node~$s(t_3)$ of Fig.~\ref{fig:Styles/NIPS_fig_1.png}.

2. \textit{Expansion}: one of those actions is randomly selected, and the child node is initialized accordingly.

3. \textit{Rollout}: starting from this child node, we select a sequence of actions uniformly at random, while accumulating the simulated instantaneous rewards along the trajectory. The resulting total is the cumulative reward associated with a leaf node. The rollout ends after a specified number of random actions (set as a hyperparameter).

4. In \textit{backpropagation}, for all nodes passed through the selection and expansion process, their number of visits is increased by one, and the cumulative rewards are updated. We repeat~$m$ simulations, similarly.

While we are at state~$s(t_0)=(\pazocal B(t_0),\pazocal G(t_0))$, we use MCTS to make the next decision. MCTS will start from root node~$s(t_0)$ and perform the simulations explained above.
Note that the~$m$ simulations occur along a ``virtual time'', i.e., they are calculated while we are at time~$t_0$ in the real system. The actions taken in such simulations have no effect in the real system: the time-expanded graph in the real system remains~$\pazocal G(t_0)$, although during simulation many other time-expanded graphs are evaluated. After MCTS terminates, we apply the most promising action in the real system. Such an action is the one corresponding to a child of the root associated with the highest simulated cumulative reward.

\subsection{Prediction and Adaptation to the Current State}
\label{sec:prediction}
Since we do not advance the actual time during the MCTS simulations, we cannot get new events from the real environment~$\textit{Env}$. However, when choosing among multiple actions, their quality is evaluated based on future rewards, which depend on future requests. We thus need a model to predict them. Such a prediction model consists of a simulated environment~$\textit{Env'}$ that generates requests during the MCTS simulations. $\textit{Env'}$ is trained on historical data, consisting of previously observed realizations~$\pazocal R_\omega$ of sequences of requests. In our extended report~\cite[\textsuperscript{\S{}\ref{sec_Requests_Generation}}]{WangExtended} we detail a prediction model trained on previous days of operations of the controlled system.\footnote{
As expected, we observed that the performance of our approach depends on the accuracy of the prediction model~(PM). We do not adopt here a sophisticated PM, which could yield overly optimistic results. We adopt instead a simple PM, to ensure a conservative evaluation. Despite this simplification, we achieve good performance. Further improvements are likely with more advanced PMs, which are beyond the scope of this work.
}

Observe that the choice of the action at time~$t_n$ is not only impacted by the predicted future requests, but also by current state~$s(t_n)=\left(\pazocal B(t_n),\pazocal G(t_n)\right)$. Within the rollout phase of MCTS, we count the number of requests that can be served, among the simulated future requests and the requests waiting in the buffer~$\pazocal B(t)$ (this buffer is the real one, not the simulated one). This mechanism thus allows the control policy to adapt to the currently waiting real requests.

\begin{algorithm}[tb]
	\caption{ Online Design of Dynamic Networks (OD\textsuperscript{2}N)  }
 \begin{footnotesize}
	\begin{algorithmic}[1]
        
	\STATE \textbf{Input} Substrate graph~$\pazocal{G}_\text{substr}=(\pazocal{V},\pazocal{E}_\text{substr})$\\
        \,\,\,\,\,\,\,\,\,\,\,\,\,\,\,\ Environment $\textit{Env}$. Lifespan $T$.\\
        \STATE \textbf{Initialization} 
        \STATE Train the auxiliary policy~$\pi_t^\text{aux}$ (\S{}\ref{sec:neural-policy})
        \STATE $n=0, t_0=0$
        \STATE Initial buffer $\pazocal B(t_0)= \emptyset$.
        \STATE \label{ln:random}
        Initialize $\pazocal{E}(t_0)$ by adding time-expanded edges randomly, while respecting Ass.~\ref{ass:sequence}, until all layers cover a period later than~$L$.
        \STATE Initial time-expanded graph $\pazocal G(t_0) = ( \pazocal{V},\pazocal{E}(t_0))$.
        \STATE Initial state~$s(t_0)= (\pazocal B(t_0), \pazocal G(t_0))$.
        \\
        \REPEAT 
        \STATE Choose action~$e_n\sim\pi(\cdot|s(t_n))$ by running MCTS (\S~\ref{sec:MCTS}). \\
        \STATE Update $\pazocal{G}({t_n})$ by adding time-expanded edge~$e_n$
        \STATE Check if any requests in the buffer~$\pazocal B(t_n^-)$ can be served, according to~\eqref{eq:change-of-B} and update the count of served requests~\eqref{eq:reward-served} accordingly.
        \STATE Calculate next intervention instant $t_{n+1}$ according to Cor.~\ref{cor:intervention-instants}.
        \REPEAT
        \FORALL{expired edge}
        \STATE Remove it from~$\pazocal G(t)$ according to case~(ii) of \S{}\ref{sec:state-transitions}.
        \ENDFOR
        \FORALL{request $d=(\tau,v,v')$ arriving in $\tau\in]t_n, t_n+1[$}
        \IF{the request can be served, i.e., $\pazocal P(\tau,v,v')\neq\emptyset$}
        \STATE Serve $d$ immediately, increment served requests count\eqref{eq:reward-served}.
        \ELSE
        \STATE Place $d$ in buffer~$\pazocal B(t_n)$
        \ENDIF
        \ENDFOR
        \UNTIL{$t\ge t_{n+1}$}
        \STATE Update state~$s(t_{n+1})$ based on the previous events\\
        \UNTIL $t_n > T$.
        \STATE \textbf{Return} Cumulated reward~$\sum_{t_n\le T} r_\omega(t_n,e_n|s(t_n))$.
	\end{algorithmic}    
\end{footnotesize}
 \label{fig:algo_1}
\end{algorithm}

\subsection{Neural Network Policy to Improve MCTS}
\label{sec:neural-policy}

In the \emph{Selection} phase of MCTS (\S{}\ref{sec:MCTS}), if the system is in state~$s=s(t_n)$, score~$\textit{uct}(s,e)$ is associated with each action~$e\in\pazocal A(t_n,s(t_n))$, and the action with the maximum score is selected. The classical score (the Upper Confidence Bound applied to trees (UCT)~\cite[\textsuperscript{\S{}2.2}]{kocsis2006bandit}) cannot be applied, since it relies on exploring a relatively large set of state-action pairs, which is infeasible in our case, due to the huge space of possible pairs, resulting from the exponential explosion of possible network trajectories. We thus need to exploit a priori knowledge to guide exploration towards interesting state-action pairs. We train an~\emph{auxiliary policy}~$\pi_t^\text{aux}(e|s)$ to capture such a priori knowledge, with~$\sum_{e\in\pazocal A(t,s(t))}\pi_t^\text{aux}(e|s)=1$ and score actions as:
\begin{align*}
\textit{uct}(s,e) = Q(s,e) + c\cdot \pi_t^\text{aux}(e|s)\sqrt{{\ln M(s)}/\left(1+M(s,e)\right)},
\end{align*}
where $Q(s,e)$ denotes the empirical cumulative reward when taking action $e$ in state $s$, $M(s)$ and $M(s,e)$ are the number of visits of state $s$ and state--action pair $(s,e)$, respectively, and $c$ is a hyperparameter which gives increased weight to exploration. Observe that~$\pi_t^\text{aux}(e|s)$ bias exploration towards pairs that are interesting based on the a-priori knowledge. This formula is similar to~\cite[\textsuperscript{(1)}]{Muller2016}, with the difference that we let~$\pi_t^\text{aux}$ depend on time, which is useful when facing a non-stationary environment (e.g., requests may follow a trend that depends on the time of day).

To capture a-priori knowledge into~$\pi^\text{aux}_t$, we create a training set $\{\pazocal{G}_1(t)_{t\in[0,T]}, \pazocal{G}_2(t)_{t\in[0,T]},\dots\}$, where each~$\pazocal{G}_z(t)_{t\in[0,T]}$ is an exemplary trajectory, i.e., evolution of a TEG over time. These trajectories may be obtained via running some rules of thumb on historical observed events, or by running, in hindsight, some heavy algorithms offline, even those that could never be run in real time \cite{Ching-AnCheng2023}. We in particular obtain such trajectories by running vanilla MTCS (with classical UCT score) with a high number~$m$ of simulations, which takes much time and can thus only be run offline. We implement~$\pi_t^\text{aux}$ as the output of a neural network and we train it via \emph{imitation learning},  so as that~$\pi_t^\text{aux}(e|s(t))$ reproduces well the probability of taking action~$e\in\pazocal A(t,s(t))$ when in state~$s(t)$, observed on the transitions of the exemplary trajectories.
The neural network is trained to maximize the likelihood of observed transitions, providing a prior $\pi^{aux}_t(e|s)$ that biases MCTS exploration. Details are in our extended report~\cite[\textsuperscript{App.~\ref{sec:neural-policy}}]{WangExtended}.

\subsection{Online Design of Dynamic Networks (OD\textsuperscript{2}N) Algorithm}

Alg.~\ref{fig:algo_1} summarizes our solution method.
It is an online method, and lifespan~$T$ can be set infinitely high.

\section{Experimental Results}
\label{sec:experimental-results}

\noindent To our knowledge, our  method is the first to address the online design of dynamic networks. We therefore compare it with domain-specific SOTA methods in 3 distinct applications: (1)~transport, (2)~configuration of a complex system, (3)~$k$-server problem. For lack of space, we discuss (1) here, while (2)-(3) are in our extended report~\cite[\textsuperscript{App.~\ref{sec:other}}]{WangExtended}. 

\subsection{Application description}
\label{eq:appplication-description}

The application presented here is the design of a futuristic public transportation system in which bus routes dynamically evolve to accommodate a sequence of unknown user trip requests. This design must be computed in real time, while the network is actively operating. 
Serving a sequence of unknown user trip requests is classically framed as a Dynamic Vehicle Routing Problem (DVRP)~\cite{rios2021recent}, where incoming requests are associated with vehicles, whose routes are adjusted on the fly. The limit of DVRP is that vehicle routes are extended individually, and no overall structure is strongly promoted. Therefore, the systems based on DVRP, such as ride-sharing or demand-responsive transport, do not efficiently consolidate demand, and each mile traveled is generally shared among just a few users. This results in a high operational cost, which is difficult to sustain for generalized everyday mobility~\cite{currie2020most}.
While in DVRP, individual vehicle routes are adapted to user requests, we invert this logic: we design a network on which users can find their path. This approach is thus similar in nature to Conventional Public Transport (CPT). Similar to CPT, by designing a structured network composed of the different lines we (i)~allow consolidating and serving high-density demand efficiently, and (ii)~enable complex paths, where users can transfer from one line to another, which is key to high system efficiency.\footnote{
Transfers are generally possible only with offline VRP~\cite{Chow2022} and not with DVRP~\cite[\textsuperscript{\S{}5.1}]{rios2021recent}. Transfers are instead natural in our dynamic network.
} 
A conceptual depiction of such elements can be found in our extended report~\cite[\textsuperscript{Fig.~\ref{fig:example}}]{WangExtended}.
There is, however, a big difference between the futuristic transport system that we design with our method and CPT: in CPT, even if lines are allowed to change over the time-of-day, such changes are entirely planned offline based on some nominal demand, and they are just executed as is, day after day. In our system, instead, lines are built online proactively, during operation, and can thus be adapted to the observed demand, even when it deviates from the nominal one.

Observe that in our proposed transport system, routes change from one day to another. To avoid user disorientation, we assume that smartphone navigation apps are available to provide passengers with the best itinerary over the current network (recall that at instant~$t$ the network schedule up to time~$t+L$ is already known, enabling path computation). The system is thus convenient for users and operators.\footnote{
\textbf{Users}. 
The willingness to use apps to navigate transport systems is shown by the increasing adoption of platforms such as Uber and Lyft and by behavioral observations (80\% of New York CPT users check their path on apps twice or more a day~\cite[\textsuperscript{p 147}]{ghahramani2016trends}). Moreover, travelers are willing to renounce to regularity and use non-regular paths, if they can arrive earlier~\cite{devarasetty2012value}.
\textbf{Operators}. Despite changing routes imposes a cognitive load on drivers, many operators already implement ``route deviations''~\cite{NationalRTAP_ADA_Toolkit_2024}. The centralized vehicle dispatching made possible by automated vehicles will further remove this concern~\cite{FagnantKockelman2015}.
}

\subsection{Formalization}
\begin{figure}[t]
\centering

\begin{subfigure}{0.2\linewidth}
    \centering
    \includegraphics[width=\linewidth]{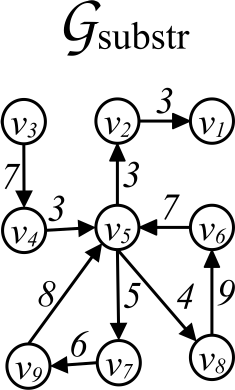}
    \label{fig:tega}
    (a)
\end{subfigure}
\hfill
\begin{subfigure}{0.55\linewidth}
    \centering
    \includegraphics[width=\linewidth]{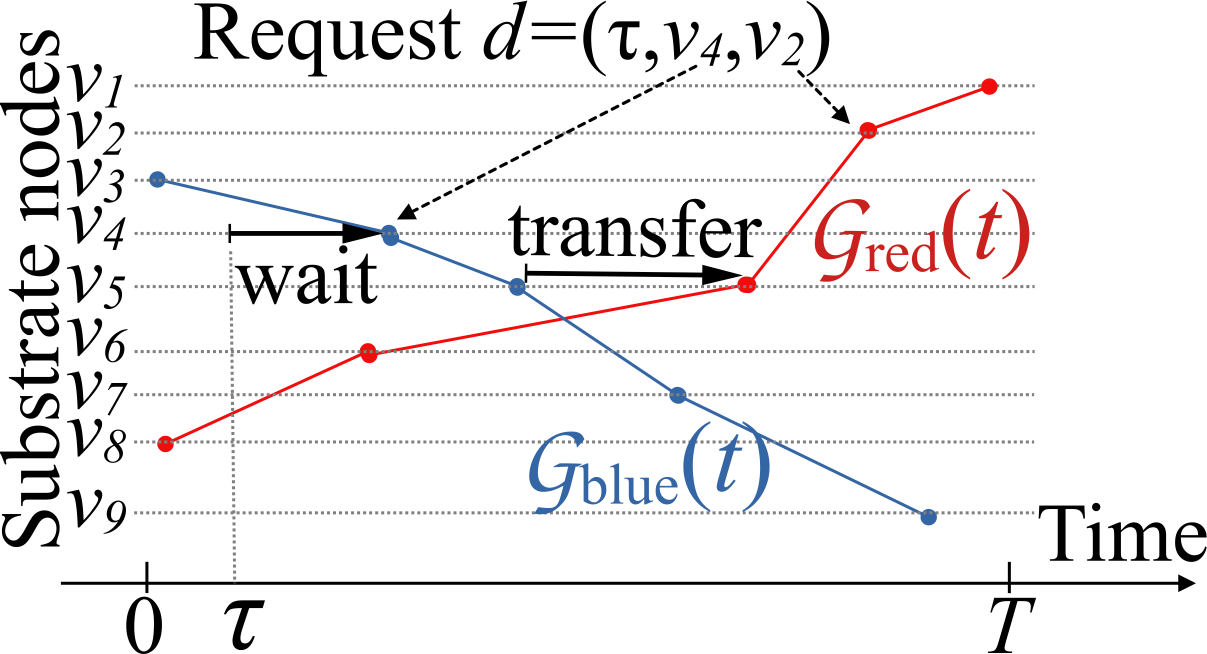}
    \label{fig:tegb}
    (b)
\end{subfigure}
\caption{On top of substrate graph~$\pazocal G_\text{substr}$ (a), we design time-expanded graph~$\pazocal G(t)$ composed of two layers (bus lines). Request~$\tau$ arrives at instant~$\tau$ and transfers from the blue to the red layer to get the service.}
\label{fig:sidebyside}
\end{figure}

In substrate graph~$\pazocal G_\text{substr}=(\pazocal V, \pazocal E_\text{substr})$, nodes~$\pazocal V$ represent potential stops, where users can board or alight, and any substrate edge~$(u,v)\in\pazocal E_\text{substr}$ represent the possibility for a vehicle to go from stop~$u$ to stop~$v$ (Fig.~\ref{fig:sidebyside}(a)), possibly traversing multiple road links (we do not represent the single road links traversed).

As shown in Fig.~\ref{fig:sidebyside}(b), each layer~$\pazocal G_l(t), l\in\pazocal L$ describes the movement of a vehicle (practically representing a transport line), i.e., $e=(\tau',u,u')\in\pazocal E_l(t)$ means that in the network planned at instant~$t$, vehicle~$l$ is scheduled to depart at instant~$\tau'$ from stop~$u$ to~$u'$. 
Environment~$\textit{Env}$ is the random process generating realizations of  user requests arriving over time, each request~$d=(\tau,v,v')\in\pazocal R_\omega$ asking a trip between stop~$v$ and~$v'$ departing no earlier than~$\tau$.
We consider each realization~$\omega\in\Omega$ as a day of operation (\S{}\ref{sec:environment-and-routing}).
Buffer~$\pazocal B(t)$ collects the requests for which no suitable path~(\eqref{eq:path-constraint-start}-\eqref{eq:max-travel}) was found until~$t$. Note that a user transfers from a vehicle to another, if their path includes multiple layers~\eqref{eq:path-continuity}.

According to Prop.~\ref{prop:timing}, we always plan the  dynamic network of bus lines~$L$ units of time in advance:
at time~$t$, all vehicle movements must be already scheduled up to time~$t+L$ (i.e., all the time-expanded edges with departure time earlier than~$t+L$ must have been added in the graph). This allows users to calculate their trip. Observe that the control actions~$\pazocal A(t,s(t))$, i.e., the insertion of new time-expanded edges in the graph, correspond to scheduling the movement of a vehicle. 

We aim to maximize the number of served requests~\eqref{eq:num_served_req} and set action rewards accordingly~\eqref{eq:reward-served} (in the other applications in our extended report~\cite[\textsuperscript{App.~\ref{sec:other}}]{WangExtended} we aim instead to minimize cost).

\subsection{Considered scenario}
\label{sec:considered-scenario}

\noindent We evaluate our method against real user requests in Manhattan (Tab.~\ref{tab:param}). We replay a randomly selected 20\% subset of requests from public dataset~\cite{TLC_Trip_Record_Data}.
We use the centroids of the 67 taxi zones as candidate pickup/dropoff stops. 
In substrate graph~$\pazocal{G}_\text{substr}$, set $\pazocal{V}$ contains such centroids, each of which is connected to all the others. Weight~$w_{u,v}$ of edge~$(u,v)\in\pazocal{E}_\text{substr}$ is the time for a bus to travel between nodes, i.e., $w_{u,v}=d(u,v)/v_\text{bus}$, where~$d_{u,v}$ is the distance between~$u,v\in\pazocal V$ and~$v_\text{bus}$ is the average speed of the bus.
The position of set~$\pazocal{L}$ of~$FS$ buses is randomly initialized.  


To train policy~$\pi_t^\text{aux}$ (\S{}\ref{sec:neural-policy}), we first construct a dataset as follows. We randomly sample 100 instances (100 days) of February 2024 from the dataset. For each sampled instance, we run a vanilla MCTS (with classical UCT scoring) between 9am: 1pm and record the resulting solution trajectory, each being a sequence of TEGs. We thus have 100 exemplary trajectories. At each decision step, we record the corresponding transition, i.e., which next stop of a certain bus is added to the trajectory. We further customize the training strategy of \S{}\ref{sec:neural-policy}, via adding into the input (i)~the ID of the stop where that bus is located before taking the action, (ii)~the travel times associated with all feasible actions (all the other stops), and (iii)~the cosine of the angle between each candidate action vector and the previous action of the same bus (we do this to let the policy learn to preserve a certain directionality of vehicle movements and avoid zig-zags). In this way, we extract movement patterns of good quality decisions from the exemplary trajectories, such as preferring next stops that are neither too close nor too far, and avoiding unnecessary detours and backtracking.

\begin{table}[t]
\centering
\begin{footnotesize}
  \begin{tabularx}{\columnwidth}{p{3.1cm}|X}
    \hline
    Parameter &Value\\
    \hline 
    Num. of candidate stops& 67\\
    Fleet size $FS$& 5, 10, 20, 30, 40, 50\\
    Speed of a bus $v_\text{bus}$
    & 17.3 km/h \cite{speed_bus_taxis}
    \\
    Private car speed $v_\text{car}$
    & 11.4 km/h \cite{speed_bus_taxis}
    \\
    Walking speed $v_\text{walk}$
    & 4.3 km/h \cite{google_map1}
    \\
    Rollout termination& 5\\
    number $N_{end}$& \\
    Look-ahead buffer~$B$& 30 mins
    \\
    Timeslots for training model~$\textit{Env}'$ (\S\ref{sec_Requests_Generation}) & 1 minute 
    \\
    Training set~$\pazocal{D}$ (\S\ref{sec_Requests_Generation})
    & All request data from~\cite{TLC_Trip_Record_Data}, February'24 
    \\
    Tested scenario (\S\ref{sec_Requests_Generation})
    & 20\% of request data from \cite{TLC_Trip_Record_Data}, 9:00-13:00 
    March 1, 2024\\
   \hline
    
    \hline
  \end{tabularx}
\end{footnotesize}
  \caption{Scenario parameters}
  \label{tab:param}
\end{table}

\begin{table}[t]
\centering
\caption{Comparison with DVRP and CPT. Hyperparameters are in \emph{italic}.}
\label{tab:comparison}
\setlength{\tabcolsep}{2pt} 
\begin{minipage}{0.48\linewidth}
\centering
Dynamic Vehicle Routing Problem (DVRP)
\begin{tabular}{|l|c|c|c|c|}
\hline
 & \textit{Max wait} & Serv. & Trip & Wait \\
 &  \textit{time (min)} &  Rate & \multicolumn{2}{c|}{time (min)} \\
\hline
\multirow{2}{*}{SOTA} & \textit{20} & 34\% & 23 & 4 \\
                      & \textit{30} & 50\% & 25 & 5 \\
\hline
\multirow{2}{*}{\shortstack{Our\\method}} & \textit{20} & 63\% & 28 & 9 \\
                                          & \textit{30} & 90\% & 31 & 14 \\
\hline
\end{tabular}
\end{minipage}
\hfill
\begin{minipage}{0.48\linewidth}
\centering
Conventional Public Transp. (CPT)
\begin{tabular}{|c|c|c|c|c|c|}
\hline
 & \textit{Lines} & Bus/ & Serv. & Trip & Wait \\
 &       & line      &  rate  & \multicolumn{2}{c|}{time (min)} \\
\hline
\multirow{4}{*}{\rotatebox{90}{SOTA}} 
& \textit{4}  & 10 & 17\% & 10 & 1  \\
& \textit{10} & 4  & 50\% & 17 & 5  \\
& \textit{20} & 2  & 73\% & 32 & 16 \\
& \textit{40} & 1  & 80\% & 56 & 41 \\
\hline
\multicolumn{3}{|c|}{\shortstack{Our method}} & 91\% & 31 & 14 \\
\hline
\end{tabular}
\end{minipage}

\end{table}

\subsection{Performance}
\label{sec:results-bus}

A good service should be capable of providing a high \textbf{service rate} (more served requests) while using few resources (small \textbf{fleet size}), and providing a good user experience, by minimizing \textbf{trip time} (average \textbf{waiting time}+in-vehicle time). 
Since our service lays in the middle between dynamic taxis and and a conventional bus network, we compare it against a SOTA DVRP method (representing the optimized dynamic shared taxi service), SOTA static bus network and the real taxi service. In all experiments, we fix default fleet size $\textit{FS}=$40 vehicles. Due to space limits, we provide here the main results. A more detailed evaluation (vehicle occupancy, sensitivity analysis) is in our extended report~\cite[\textsuperscript{\S{}\ref{sec:detailed-performance-results}}]{WangExtended}.



\subsubsection{Comparison with State-Of-The-Art (SOTA) Dynamic Vehicle Routing Problem~(DVRP) solutions.}


In Table~\ref{tab:comparison}, we compare our solutions with a well-known SOTA DVRP solution~\cite{alonso2017demand}, of which we run an open source implementation~\cite{Ride_Sharing}, with two different values of hyperparameter ``maximum waiting time'' and fleet size of 40 vehicles for both solutions and infinite seat capacity.\footnote{
We verify in our extended report~\cite[Tab.\ref{tab:3}]{WangExtended} that occupancy is compatible with real-world vehicle capacities.
} 
Observe that the goal of SOTA~\cite{alonso2017demand} is to optimize an on-demand ``high-capacity'' ride-sharing. However, the DVRP-based approach limits the actual capacity they can sustain, whereas, via our network-centric approach, we nearly double the service rate, thanks to the efficiency brought by the network structure we design online. The conceptual reasons of such gains, explained in \S{}\ref{eq:appplication-description}, are confirmed by these results.\footnote{
Observe that both the SOTA DVRP solution and our method are applied on the same demand and substrate network~$\pazocal G_\text{substr}$ (\S{}\ref{sec:Time-expanded-graph}). What changes between the two is the way vehicle movement is scheduled on top of~$\pazocal G_\text{substr}$: the former modifies vehicle routes at every request, while we design a ``masterplan'' time-expanded network~$\pazocal G(t)$ (overlay on top of~$\pazocal G_\text{substr}$), which vehicles will follow.
}
On the other hand, a loss is experienced by users (waiting and trip time, transfers), but it is acceptable, since it is compatible with what they currently experience with Conventional Public Transport~(CPT).\footnote{
\label{fn:vs-taxis}
We do not aim to replace taxis, which would still be preferred by users willing to pay a high fee for maximum comfort; our system is rather a complement to CPT for everyday mobility. In other words, our method enables a brand-new transport service, rather than a ``better'' taxi service. Such a new service sits in the middle between the efficiency and high capacity of CPT and the adaptivity and user-comfort of taxis.
}

\subsubsection{Comparison with SOTA static bus network design} \cite{fielbaum2024design}, representing conventional public transport (CPT), under the same training and test set for both (Table~\ref{tab:param}), and the same fleet size$=40$. Table~\ref{tab:comparison} shows that by increasing the number of lines, the static bus system covers more areas and improves the service rate. However, some areas remain always unreachable, leading to 20\% unserved demand. Moreover, when increasing the number of lines, trip time and waiting time increase, since fewer vehicles operate on each line. In contrast, our online design offers much higher service rates without excessively degrading trip and waiting times. The reason of this superiority was explained in~\S{}\ref{eq:appplication-description}.

\subsubsection{Comparison with real taxis}
From~\cite{TLC_Trip_Record_Data1} we measure that taxi real trips take in real-world $\sim 13$ minutes on average between 9am and 1pm, which are shorter than the ones our method provides (Table~\ref{tab:comparison}). However, we use only 40 buses, versus 2600 taxis and serve 90\% of the demand.\footnote{
Around 13k taxis operate in Manhattan~\cite{TLC_Trip_Record_Data1} and 2.6k correspond to the 20\% of 13k (remember that we considered 20\% of the real demand~Tab.~\ref{tab:param}).
} The loss of user-centric performance is thus justified by the much greater efficiency, which is compatible with the nature of the proposed system (see footnote~\ref{fn:vs-taxis}).

\subsubsection{Service characteristics}

\begin{figure}[]
  \centering
  \includegraphics[width=0.4\textwidth]{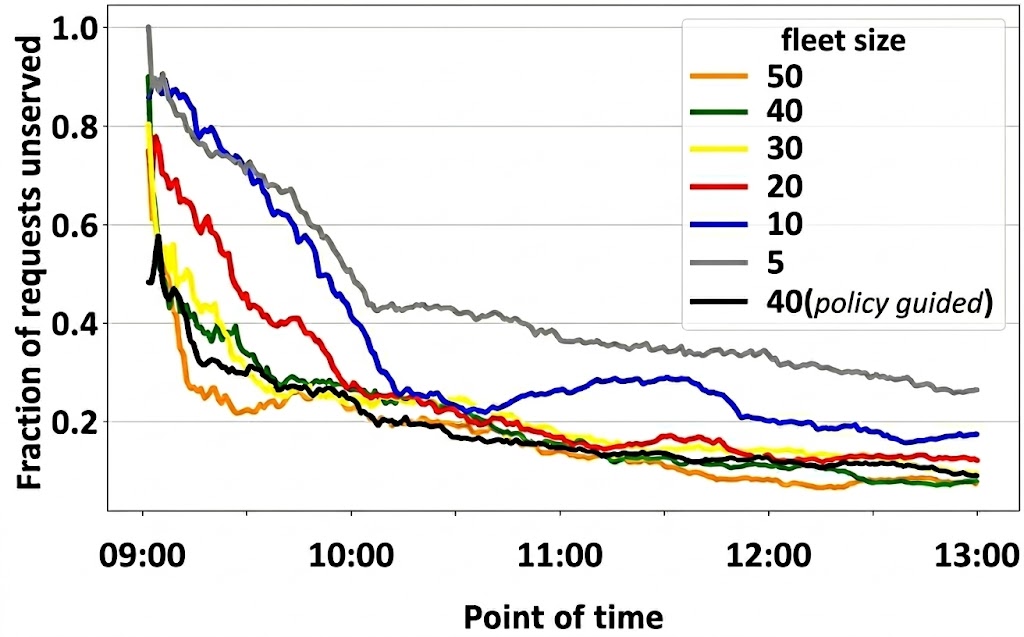}
  \caption{Fraction of requests unserved from 9:00 to 13:00.}
  \label{fig:3-1}
\end{figure}

\begin{figure}[]
  \centering
  \includegraphics[width=1.0\linewidth]{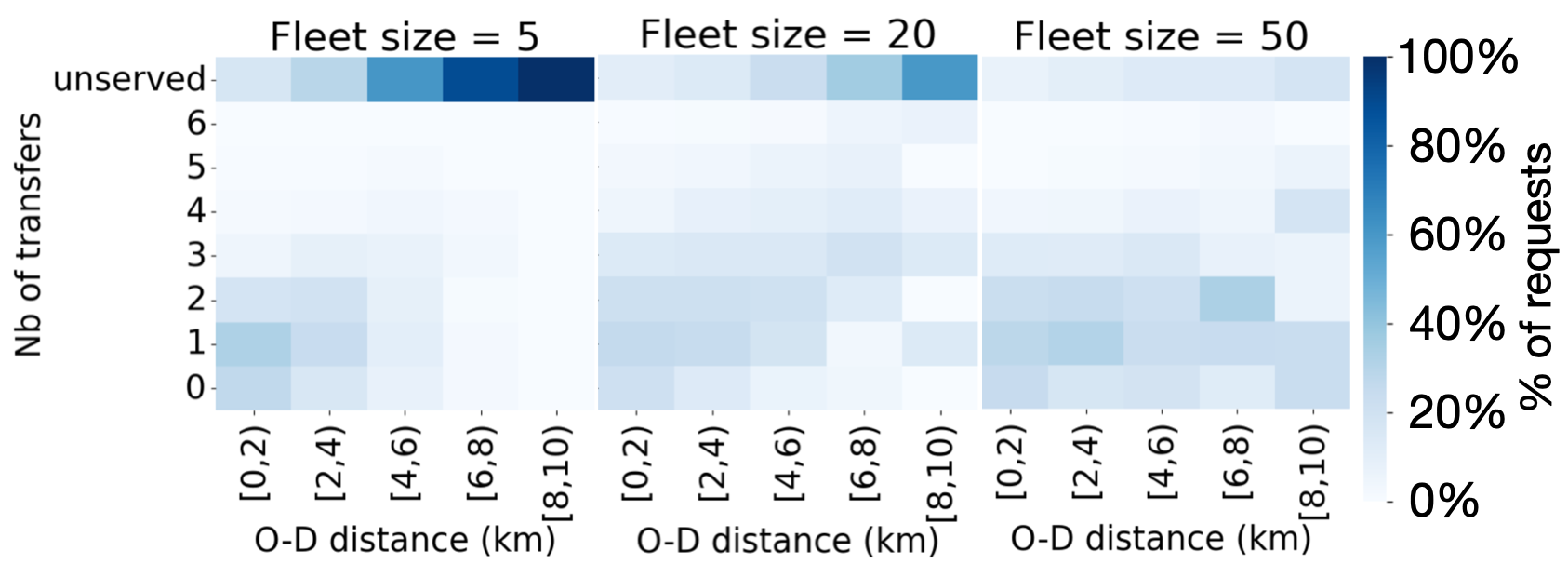}
  \caption{ Fraction of requests experiencing certain number of transfers over total requests, per each O-D distance interval. }
  \label{fig:heat_map_OD_distance_transfer}
\end{figure}

Fig.~\ref{fig:3-1} shows that the neural policy guidance is beneficial at the beginning, as it suggests directly good decisions, thus reducing from 40\% to 30\% unserved requests 30 minutes after startup, with fleet size~$\textit{FS}=40$. Fig.~\ref{fig:heat_map_OD_distance_transfer} shows that increasing fleet size allows serving longer trips. To adapt our approach to the case in which users accept no more than 2 transfers, we can assume infeasible any more complex path. In our current system, this would preserve 88\% of completed requests with $\textit{FS=}$40.

\subsubsection{Real-time computation}
We pinpoint that our method meets the real-time requirements, since the choice of each action is made within seconds, i.e., much faster than the time between two consecutive intervention instants (Cor.~\ref{cor:intervention-instants}).

In summary, the presented transport service offers a higher service rate with a smaller fleet size without excessively degrading trip time.

\section{Conclusion}


We introduced the novel problem of designing dynamic networks online, to adapt to a stochastic environment and adjust their evolution to maintain satisfying performance. We formalized a general online decision process and a Monte Carlo Tree Search-based solution method, guided by a neural network policy.
The numerical results show that, by operating real systems via designing, online, their network structure, brings important performance improvements.

\section{Acknowledgment}
\begin{scriptsize}Supported by the French ANR research project MuTAS (ANR-21-CE22-0025-01) \end{scriptsize}

\bibliographystyle{IEEEtran}
\bibliography{refs}


\newif\ifshowappendix
\showappendixtrue  
\ifshowappendix
\clearpage

\begin{appendices}

\section{Table of notation}
Table~\ref{tab:notation} summarizes the notation used in this paper.
\todo{to be checked}

\begin{table*}[h!]
\centering
\caption{Notation}
\label{tab:notation}
\setlength{\tabcolsep}{4pt}
\renewcommand{\arraystretch}{1.05}
\begin{tabular}{|l|p{7.0cm}|p{4.2cm}|}
\hline
\textbf{Symbol} & \textbf{Meaning} & \textbf{First introduction} \\
\hline
\multicolumn{3}{|c|}{\textbf{Sets, graphs, and spaces}} \\
\hline
$\pazocal V$ & Set of substrate nodes & Sec.~\ref{sec:Time-expanded-graph} \\
\hline
$\pazocal E_{\text{substr}}$ & Set of substrate edges & Sec.~\ref{sec:Time-expanded-graph} \\
\hline
$\pazocal G_{\text{substr}}$ & Substrate graph & Sec.~\ref{sec:Time-expanded-graph} \\
\hline
$\pazocal L$ & Set of layers & Eq.~\eqref{eq:multiple-layers} \\
\hline
$\pazocal G_l(t)$ & Time-expanded graph of layer $l$ at time $t$ & Eq.~\eqref{eq:multiple-layers} \\
\hline
$\pazocal G(t)=(\pazocal V,\pazocal E(t))$ & Overlay time-expanded graph at time $t$ & Sec.~\ref{sec:Time-expanded-graph} \\
\hline
$\pazocal E_l(t)$ & Set of active time-expanded edges in layer $l\in L \pazocal L$ at time $t$ & Eq.~\eqref{eq:multiple-layers} \\
\hline
$\pazocal E(t)=\bigcup_{l\in\pazocal L} \pazocal E_l(t)$ & Set of active time-expanded edges at time $t$ & Sec.~\ref{sec:Time-expanded-graph} \\
\hline
$\pazocal R_\omega$ & Set of requests in realization $\omega\in\Omega$ & Sec.~\ref{sec:environment-and-routing} \\
\hline
$\Omega$ & Set of realizations of the environment & Sec.~\ref{sec:environment-and-routing} \\
\hline
$\pazocal P(v,v',t,\pazocal G)$ & Set of feasible paths in graph $\pazocal G(t)$ from $v$ to $v'$
& Sec.~\ref{sec:environment-and-routing}, before \eqref{eq:arrival-time} \\
\hline
$\pazocal A(t,s)$ & Set of admissible control actions at time $t$ in state $s$ & Sec.~\ref{sec:control} \\
\hline
\multicolumn{3}{|c|}{\textbf{State, events, and decisions}} \\
\hline
$t$ & Time variable & Throughout; see Sec.~\ref{sec:model} \\
\hline
$e=(\tau,u,u')$ & Time-expanded edge departing at time $\tau$ from $u$ to $u'$ & Sec.~\ref{sec:Time-expanded-graph} \\
\hline
$d=(\tau,v,v')$ & Request arriving at instant~$\tau$ (also corresponding to the earliest departure time), origin $v$, and destination $v'$ & Sec.~\ref{sec:environment-and-routing} \\
\hline
$\pazocal B(t)$ & Buffer of pending requests at time $t$ & Sec.~\ref{sec:state-transitions} \\
\hline
$s(t)=\big(\pazocal B(t),\pazocal G(t)\big)$ & State at time $t$ & Sec.~\ref{sec:state-transitions} \\
\hline
$s(t^-),\,s(t^+)$ & State immediately before / after an intervention or jump at time $t$ & Used in proofs, App.~\ref{sec:proofs} \\
\hline
$\tau_n$ & Departure time of the $n$-th inserted time-expanded edge & Sec.~\ref{sec:control} \\
\hline
$t_n$ & Intervention instant associated with the $n$-th action & Prop.~\ref{prop:timing}, Cor.~\ref{cor:intervention-instants} \\
\hline
$e_n$ & Time-expanded edge inserted at intervention instant $t_n$ & Sec.~\ref{sec:control} \\
\hline
$ICS=(t_n,e_n)_{n\ge 0}$ & Impulsive control strategy & Sec.~\ref{sec:control} \\
\hline
$\pi$ & Policy in the MDP formulation & Sec.~\ref{sec:mdp} \\
\hline
$\pi_\omega$ & Realization of policy $\pi$ under $\omega\in\Omega$ & Sec.~\ref{sec:mdp} \\
\hline
$\pi_t^{\text{aux}}(e\mid s)$ & Auxiliary policy guiding MCTS exploration & Sec.~\ref{sec:neural-policy} \\
\hline
\multicolumn{3}{|c|}{\textbf{Parameters, functions, and performance metrics}} \\
\hline
$w_{u,u'}$ & Weight / traversal time of substrate edge $(u,u')$ & Sec.~\ref{sec:Time-expanded-graph} \\
\hline
$c_{u,u'},c_{u,u',\omega}$ & Cost associated with substrate edge $(u,u')$ and its realization~$\omega\in\Omega$ & Sec.~\ref{sec:Time-expanded-graph} \\
\hline
$L$ & Maximum path duration / decision lag & Eq.~\eqref{eq:max-travel}; Prop.~\ref{prop:timing} \\
\hline
$N^\pi_\omega$ & Number of served requests under policy realization $\pi_\omega$ & Eq.~\eqref{eq:num_served_req} \\
\hline
$C^\pi_\omega$ & Cumulative cost under policy realization $\pi_\omega$ & Eq.~\eqref{eq:cumulative-cost} \\
\hline
$r_\omega(t_n,e_n\mid s(t_n))$ & Instantaneous reward at intervention instant $t_n$ & Eqs.~\eqref{eq:reward-served}--\eqref{eq:reward-cost} \\
\hline
$\textit{uct}(s,e)$ & Tree-search score used in MCTS selection & Sec.~\ref{sec:neural-policy} \\
\hline
$Q(s,e)$ & Empirical mean return of action $e$ in state $s$ & Sec.~\ref{sec:neural-policy} \\
\hline
$M(s)$ & Number of visits of state $s$ in MCTS & Sec.~\ref{sec:neural-policy} \\
\hline
$M(s,e)$ & Number of visits of state--action pair $(s,e)$ in MCTS & Sec.~\ref{sec:neural-policy} \\
\hline
$T$ & Final time / time horizon & Sec.~\ref{sec:mdp} and Alg.~\ref{fig:algo_1} \\
\hline
\end{tabular}
\end{table*}

\section{Proofs}
\label{sec:proofs}
\subsection{Proof of Proposition~\ref{prop:PDMP}}
\begin{proof}
In the absence of control actions, the state changes only due to two types of events: exogenous arrivals of requests generated by the environment and placed in the buffer (case~(i) of \S{}\ref{sec:state-transitions}) and deterministic removals of time-expanded edges whose departure times have been reached (case~(ii) of \S{}\ref{sec:state-transitions}). Let us call \emph{event times} the instants at which these two types of event occur.

Between two consecutive event times, there is no change in state~$s(t)=(\pazocal B(t),\pazocal G(t))$ and all the requests that arrive are directly routed (case~(a) of \S{}\ref{sec:state-transitions}), inducing no modification of the buffer, and thus no modification of the state. Hence the state remains unchanged, so its evolution is deterministic between jumps.

At the time when a request is added to buffer~$\pazocal B(t)$, the state undergoes an instantaneous jump caused by the update of the buffer. The occurrence of this jump is stochastic and depends jointly on the exogenous arrival process and the state (in particular current graph~$\pazocal G(t)$, determining whether the arrived request can be directly routed or must be placed in the buffer).

At the instant in which the departure time of an edge currently present in the graph is reached, that edge is removed, which also induces an instantaneous jump of the state, caused by the update of time-expanded graph~$\pazocal G(t)$. Since such removals are fully determined by the current state $s(t)$ (and in particular time-expanded graph~$\pazocal G(t)$), their occurrence is deterministic given $s(t)$.

By assumption, future arrivals are exogenous and independent of the past history and of the current state. Therefore, conditional on the present state $s(t)$, the law of the future state evolution depends only on $s(t)$ and not on the past trajectory. Hence the process is Markov.

Since the process is Markov, evolves deterministically between event times, and undergoes instantaneous jumps at those event times, it is a piecewise deterministic Markov process.
\end{proof}

\subsection{Proof of Proposition~\ref{prop:timing}}
\begin{proof}
Let us take any impulsive control strategy~$\textit{ICS}=\left(t_n, e_n=(\tau_n,v_n,v_n')\right)_{n\ge 0}$ and any realization~$\omega\in\Omega$. Let us consider strategy~$\textit{ICS}_{t_n'}$, which is identical to~$\textit{ICS}$, except for the instant in which time-expanded edge~$e_n$ is inserted: this instant is~$t_n'$ instead of~$t_n$. Observe that~$\textit{ICS}_{t_n'}=\textit{ICS}$, if~$t_n'=t_n$. Let~$\pazocal P_{t_n'}(v,v,t)$ denote the pathset available in the time-expanded graph at time~$t$ when applying strategy~$\textit{ICS}_{t_n'}$.

The number of served requests~\eqref{eq:num_served_req} when applying strategy~$\textit{ICS}_{t_n'}$ can be written as:
\begin{align}
N_\omega^{\textit{ICS}_{t_n'}}
=
\sum_{d=(\tau,v,v')\in\pazocal R_\omega}
&\mathbbm 1\big\{
\tau\in[0,T], 
\nonumber
\\
&\exists t\in]\tau,T[:\pazocal P_{t_n'}(v,v',t)\neq 0 \big\}.
\label{eq:N}
\end{align}
\underline{PART A}) \emph{Benefits of anticipation}

We will show that, for any $(v,v')\in\pazocal E, t\ge 0$, we have
\begin{align}
\label{eq:implication1}
t_n' < t_n 
&\Rightarrow \pazocal P_{t_n'}(v,v',t) \supseteq \pazocal P_{t_n}(v,v',t),\forall t\ge 0
\\
\label{eq:implication2}
&\Rightarrow N_\omega^{\textit{ICS}_{t_n'}} \ge N_\omega^\textit{ICS}.
\end{align}
 
To understand implication~\eqref{eq:implication1}, let us fix any instant~$t\ge 0$, and let~$\pazocal E_{t_n}(t)$ and~$\pazocal E_{t_n'}(t)$ denote the set of time-expanded edges active at time~$t$ under strategy~$\textit{ICS}$ and~$\textit{ICS}_{t_n'}$ respectively. 

CASE 1) If~$t<t_n'<t_n$, then~$\pazocal E_{t_n}(t)=\pazocal E_{t_n'}(t)$ and thus~$\pazocal P_{t_n'}(v,v',t) = \pazocal P_{t_n}(v,v',t)$. 

CASE 2) The same reasoning holds when~$t_n'<t_n \le t$. 

CASE 3) Let us now consider the case~$t\in [t_n',t_n[$. In such~$t$, we have~$\pazocal E_{t_n'}(t)=\pazocal E_{t_n}(t)\cup\{e_n\}$, since in such instants~$t$ time-expanded edge~$e_n$ would be be already present with strategy~$\textit{ICS}_ {t_n'}$ and absent with strategy~$\textit{ICS}$. At such instants, $\pazocal P_{t_n'}(v,v',t) \supsetneq \pazocal P_{t_n}(v,v',t)$. Therefore, in general, $\pazocal P_{t_n'}(v,v',t) \supseteq \pazocal P_{t_n}(v,v',t),\forall t\ge 0$. Implication~\eqref{eq:implication2} is a simple consequence of~\eqref{eq:N}.

Equations~\eqref{eq:implication1}-\eqref{eq:implication2} show that it is preferable to add time-expanded edges as soon as possible. However, too much anticipation in the insertion of time-expanded edges is not needed. 

\underline{PART B}) \emph{Too much anticipation is not needed.}

More precisely, we will now show that inserting time-expanded edge~$e_n=(\tau_n,v_n,v_n')$ before~$\tau_n-L$ brings the same outcome in terms of~\eqref{eq:N}. Let us assume~$t_n<\tau_n-L$. Let us take any request~$d=(\tau,u,u')\in\pazocal T_\omega$ that is served under strategy~$\textit{ICS}=\textit{ICS}_{t_n}$, i.e., 
$$\mathbbm 1\big\{ \tau\in[0,T], \exists t\in]\tau,T[:\pazocal P_{t_n}(v,v',t)\neq 0 \big\}=1.$$
Let us assume that request~$d$ uses a path that does not contain time-expanded edge~$e_n$. Then, inserting~$e_n$ at time~$\tau_n-L>t_n$ does not change anything and still~$\mathbbm 1\big\{ \tau\in[0,T], \exists t\in]\tau,T[:\pazocal P_{\tau_n-L}(v,v',t)\neq 0 \big\}=1$. Suppose now that, with strategy~$\textit{ICS}$, such request~$d$ uses path~$p\in\pazocal P_{t_n}(v,v',t)$ that contains~$e_n$. By constraints~\eqref{eq:path-constraint-start}-\eqref{eq:max-travel}, we have that
$$
\tau_n-t < AT(p)-t \le L \Longrightarrow t > \tau_n-L.
$$
Therefore, inserting edge~$e_n$ at time~$\tau_n-L>t_n$ still makes path~$p$ available at instant~$t$, i.e., $p\in\pazocal P_{\tau_n-L}(v,v',t)$. Therefore
$$\mathbbm 1\big\{ \tau\in[0,T], \exists t\in]\tau,T[:\pazocal P_{\tau_n-L}(v,v',t)\neq 0 \big\}=1,$$
and thus $N_\omega^{\textit{ICS}_{\tau_n-L}} \ge N_\omega^\textit{ICS}$.
\end{proof}

\subsection{Lemma on the intervention instants}
The lemma below is needed to prove Cor.~\ref{cor:intervention-instants}. Before stating it, let us observe that~Ass.~\ref{ass:sequence} implies that
\begin{align}
\pazocal G_l&(t)
=(\pazocal V, \pazocal E_l(t)); \,\,\,\,\,
\pazocal E_l(t) = \left \lbrace 
    e_l^{[i]}=(\tau_l^{[i]},v_l^{[i]}, v_l^{{[i]}\prime})    
\right \rbrace_{i=0,1,\dots}
\nonumber
\\
& 
v^{[i]\prime} = v^{[i+1]}
\text{ and }
\tau_l^{[i+1]}=\tau_l^{[i]}+w_{v_l^{[i]},v^{[i]\prime}}, \forall l\in\pazocal L
\label{eq:connectivity}
\end{align}

\begin{lemma}
\label{lem:intervention-instants}
For any~$t\in [0,t]$, let us define
\begin{align}
e_l^\text{last}(t)=(\tau_l^\text{last},v_l^\text{last},v_l^{\text{last}\prime})
&:=
\argsup\limits_{e=(\tau,u,u')\in\pazocal E_l(t)}  \tau+w_{u,u'}
\nonumber
\\
z_l^\leftarrow(t)
&:
=\tau_l^\text{last} + w_{v_l^\text{last},v_l^{\text{last}\prime} },
\,\,\,\, \forall l\in\pazocal L
\nonumber
\\
\pazocal L^\text{inc}(t)
&
:= \{l \in \pazocal L | z_l^\rightarrow(t)<T \},
\end{align}
which denote the last edge of layer~$l$, the last instant covered by~$l\in\pazocal L$ in graph~$\pazocal G(t)$ and the set of lines that are ``incomplete'', i.e., they do not cover the entire lifespan~$[0,T]$.
If a strategy complies with Prop.~\ref{prop:timing} and constraint~\eqref{eq:connectivity}, the intervention instants are determined as follows:
\begin{align*}
t_{n+1} =
\inf_{l\in\pazocal L^\text{inc}(t_n)} z_l^\leftarrow(t_n) \text{ if }\pazocal L^\text{inc}(t_n)\neq \emptyset.
\end{align*}
If all lines are complete~$\pazocal L^\text{inc}= \emptyset$, the control strategy terminates.
\end{lemma}

\begin{proof}
Consider any impulsive control strategy~$\textit{ICS}=(t_n,e_n=(\tau_n, v_n, v_n'))_{n\ge 0}$ and any intervention~$n\ge 0$. In compliance with Prop.~\ref{prop:timing}, $t_n=\tau_n-L$. 
If~$\pazocal L^\text{inc}(t_n)=\emptyset$, the design terminates, according to~\eqref{eq:lifespan}. Let us instead assume that~$\pazocal L^\text{inc}(t_n)\neq\emptyset$.

Let~$l \in \pazocal L^\text{inc}(t_n)\neq\emptyset$ be the layer in which edge~$e_n$ was added, i.e., $e_n\in\pazocal E_l(t_n)$. To match constraint~\eqref{eq:connectivity}, we need to ``connect''~$e_n$ to the last edge~$e_l^\text{last}=(\tau_l^\text{last},v_l^\text{last},v_l^{\text{last}\prime})\in\pazocal E_l(t_n^-)$. This edge was inserted in some instant~$t<t_n$ and, due to Prop.~\ref{prop:timing} $\tau_l^\text{last}=t+L$ and thus
$$\boxed{\tau_n+w_{v_n,v_n'}}=t_n+L+w_{v_n,v_n'}
\boxed{>}
t+L+w_{v_n,v_n'}\ge t+L = 
\boxed{\tau}.$$ 
Therefore, the arrival node and instant of edge~$e_n$ cannot be connected to the departure node and instant of the last edge~$e_l^\text{last}$ in~$l$. The only possibility that the insertion of~$e_n$ does not violate constraint~\eqref{eq:connectivity} is that its departure node and instant corresponds to the arrival node and instant of~$e_l^\text{last}$, i.e., 
\begin{align}
\label{eq:attachment}
\tau_l^\text{last}+w_{v_l^\text{last},v_l^{\text{last}\prime}}=\tau_n \text{ and } v_l^{\text{last}\prime}=v_n. 
\end{align}

Let us assume by contradiction that
\begin{align}
\tau_n
& \neq 
\inf_{l\in\pazocal L}
\left[
\sup\{\tau+w_{u,u'}| e=(\tau,u,u')\in\pazocal E_l(t_n)\}
\right]
\nonumber
\\
& 
\Rightarrow
\exists l'\in\pazocal L:
\tau_{l'}^\text{last}+w_{v_{l'}^\text{last},v_{l'}^{\text{last}\prime}}
<
\tau_l^\text{last}+w_{v_l^\text{last},v_l^{\text{last}\prime}}
\\
& 
\Rightarrow^\eqref{eq:attachment}
\tau_n >
\tau_{l'}^\text{last}+w_{v_{l'}^\text{last},v_{l'}^{\text{last}\prime}}
\\
& 
\Rightarrow^{\text{Rem.}~\ref{rem:restriction}}
t_n + L >
\tau_{l'}^\text{last}+w_{v_{l'}^\text{last},v_{l'}^{\text{last}\prime}}
\label{eq:last-attach}
\end{align}

This would mean that line~$l'$ would remain forever incomplete: at any other future instant~$t>t_n$, according to Rem.~\ref{rem:restriction} we would only add edges whose departure time is~$t+L>t_n+L >^\eqref{eq:last-attach} \tau_{l'}^\text{last}+w_{v_{l'}^\text{last},v_{l'}^{\text{last}\prime}}$, which cannot be added to~$l'$ (to comply with Prop.~\ref{eq:connectivity}). This would violate~\eqref{eq:lifespan}, which is absurd.
\end{proof}

\subsection{Proof of Corollary~\ref{cor:intervention-instants}}
\begin{proof}
The function~$f^\text{next}(\cdot)$ is the one stated in Lemma~\ref{lem:intervention-instants}.
\end{proof}

\subsection{Proof of Proposition~\ref{prop:objective}}

\begin{proof}
We consider the two possible definitions of the instantaneous reward separately.

\paragraph{Case 1: reward defined by~\eqref{eq:reward-served}}
Let us fix policy~$\pi$, and let~$\omega\in\Omega$ be a realization. Consider a request
$d=(\tau,v,v')\in \pazocal R_\omega$ with $\tau\in[0,T]$ that is eventually served under~$\pi$.
The time at which~$d$ is served is
\[
st(d)=\inf\{t\ge \tau \mid \pazocal P(v,v',t,\pazocal G(t))\neq \emptyset\}.
\]
Hence there exists a unique intervention index~$n$ such that either:

\begin{itemize}
    \item $t_n<\tau<t_{n+1}$ and the request becomes feasible during the interval
    $[\tau,t_{n+1}]$, in which case it is counted by the first term of
    \eqref{eq:reward-served} at instant~$t_n$; or
    \item the request is already in the buffer at time~$t_n$ and becomes feasible at~$t_n$,
    in which case it is counted by the second term of~\eqref{eq:reward-served}.
\end{itemize}

In both cases the request contributes exactly once to the sum
$\sum_{t_n\le T} r_\omega(t_n,e_n\mid s(t_n))$.
Indeed, a request cannot be counted twice: once a request is served, it is no longer kept in the
buffer, and thus it cannot contribute again at a later intervention instant.
Therefore, for every realization~$\omega$ and every policy~$\pi$, if let~$\textit{ICS}_{\pi,\omega}$ denote the impulsive control strategy induced by~$\pi$ (according to \S{}\ref{sec:mdp}) and~$N_\omega^\pi$ the number of served requests in the interval~$[0,T]$ (according to~\eqref{eq:N-omega-pi}), we get
\[
\sum_{t_n\le T} r_\omega(t_n,e_n\mid s(t_n))
=
N_\omega^\pi,
\]
Taking expectations yields
\[
\mathbb E_\omega\!\left[\sum_{t_n\le T} r_\omega(t_n,e_n\mid s(t_n))\right]
=
\mathbb E_\omega[N_\omega^\pi]
\overset{\eqref{eq:N-pi}}{=}
N^\pi
.
\]
Hence maximizing objective~\eqref{eq:max} is equivalent to maximizing the expected number of
served requests.

\paragraph{Case 2: reward defined by~\eqref{eq:reward-cost}}
In this case, for every realization~$\omega$,
\[
\sum_{t_n\le T} r_\omega(t_n,e_n\mid s(t_n))
=
-\sum_{t_n\le T} c_{v_n,v_n',\omega}
\overset{\eqref{eq:cumulative-cost}}{=}
-C_\omega^{\textit{ICS}_{\pi_\omega}}
\overset{\eqref{eq:N-pi}}{=}
-C_\omega^\pi
.
\]
By taking the expectation:
\[
\mathbb E_\omega\!\left[\sum_{t_n\le T} r_\omega(t_n,e_n\mid s(t_n))\right]
=
-\mathbb E_\omega[C_\omega^\pi],
\]
where $C_\omega^\pi$ denotes the cumulative cost under policy~$\pi$.
Thus maximizing objective~\eqref{eq:max} is equivalent to minimizing the expected cumulative cost.

Combining the two cases proves the claim.
\end{proof}

\section{Detailed analysis of the transport application}
\label{sec:oana}

 \begin{figure}
   \centering
   \includegraphics[width=0.9\linewidth]{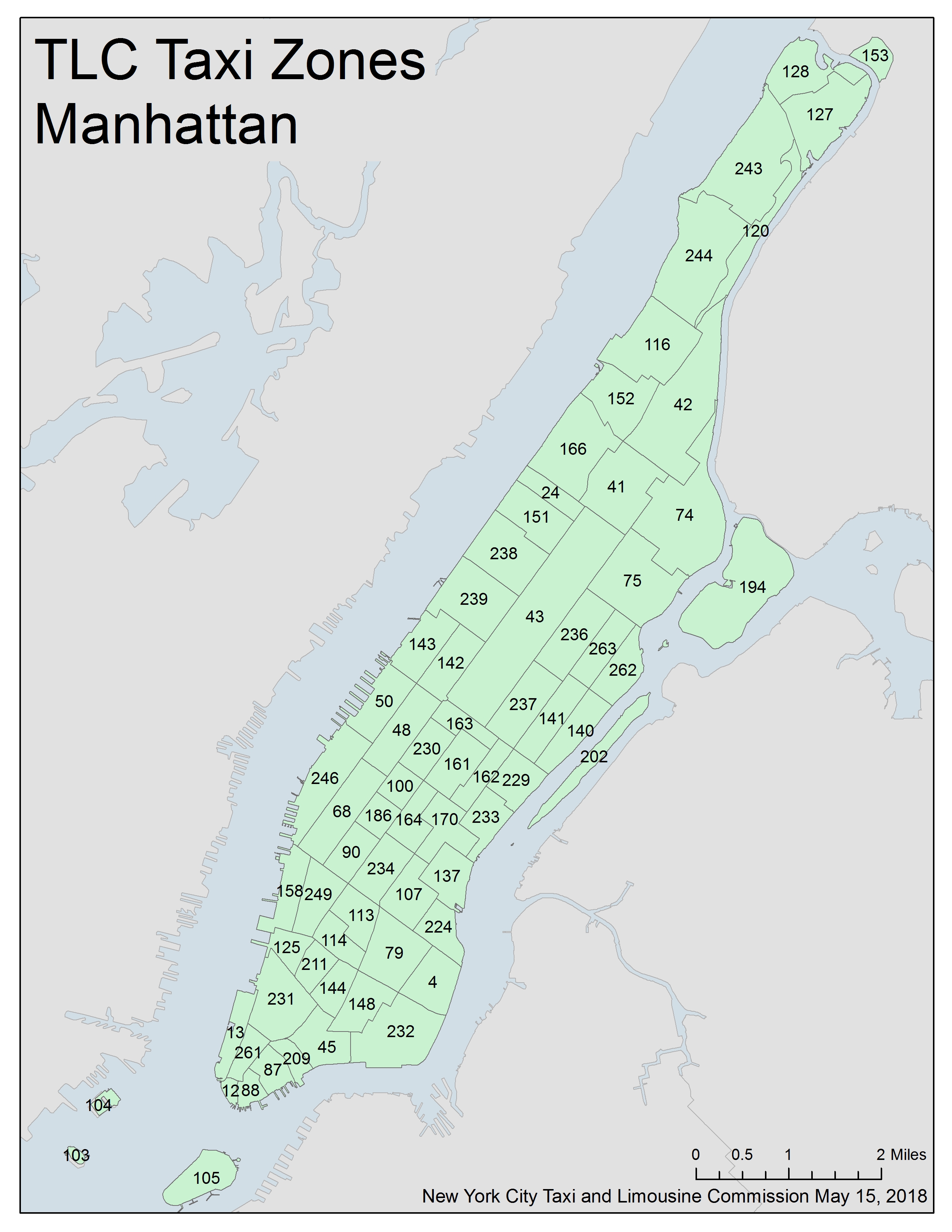}
   \caption{Taxi zones in Manhattan.}
   \label{fig:taxi_zone_map_manhattan}
 \end{figure}

We evaluate the performance of our method on a dataset of real trip requests in Manhattan, as described in \S{}\ref{sec:considered-scenario}, where we consider the centroids of the taxi zones (depicted in Fig.~\ref{fig:taxi_zone_map_manhattan}) as potential stops. Each taxi zone is on average 0.88 Km\textsuperscript{2}. The radius of an equivalent circle is 530 m, which would require 7.4 minutes to walk. If we assume that users origins and destinations can be anywhere in Manhattan, it implies that a user walks on average 7.4 minutes to go from their origin location to the closest stop or from the last stop of their trip to their final location.

\subsection{Requests Generation Model}
\label{sec_Requests_Generation}

As explained in \S\ref{sec:MCTS}, Monte-Carlo Tree Search (MCTS) requires simulating future possible evolution trajectories of the system, so that one can choose the next action that will likely induce the most desirable evolution. To perform such simulation, we need a model~$\textit{Env'}$ of the environment, which generates events (requests in this case) that are statistically similar to those generated by the real environment~$\textit{Env}$. We can fit~$\textit{Env}'$ on historical data of real-world environment~$\textit{Env}$. In particular, in the transport case we are tackling here, model~$\textit{Env}'$ can be trained on historical dataset~$\pazocal D$ of previously observed trips.
Requests are a spatial-temporal marked point process~\cite{Handcock1994}, which is a type of marked point process mentioned in \S{}\ref{sec:environment-and-routing}. Therefore, $\textit{Env}'$ must be able to mimic the spatial patterns (where origins and destinations of trips are distributed) and temporal patterns (rate of requests). 

\emph{1. Learning temporal patterns}. 
For each day in the past, we first count the number of requests generated at different timeslots. We train a support-vector machine (SVM) to predict the request counts of all timeslots in the next day.

\emph{2. Learning spatial patterns}.  We count the origin and destination of all past requests and build an OD matrix~$M = \{a_{uv}\}_{u,v\in\pazocal V}$ where~$a_{uv}$ is the fraction of observed trips that went from origin~$u$ to destination~$v$. The value of $a_{uv}$ can be interpreted as the probability that, given any trip request, its origin will be~$u$ and its destination~$v$.

When, within MCTS, we use $\textit{Env}'$ to generate simulated requests, we do as follows. We first predict the number of future requests via the SVM at each timeslot.
For each of them, we randomly select the origin and the destination based on probabilities~$a_{u,v}$.
Note that we deliberately adopted this simple prediction model~$\textit{Env}'$, so as to provide a conservative indication of the performance of the proposed system. Indeed, already with this simple model, \S{}\ref{sec:results-bus} shows that the proposed method outperforms SOTA DVRP methods. One could than replace this simple~$\textit{Env}'$ with any of the advanced demand prediction models available in the literature, and expect even better performance. This would however fall outside the scope of this work. 




\begin{table}[t]
\centering
\begin{small}
  \begin{tabularx}{\columnwidth}{p{3.8cm}|X}
    \hline
    Number of layers of MLP         & 3                           \\
    Hidden dimension             & 64                          \\
    Activation function          & ReLU                        \\
    Loss function                & Negative log-probability loss \\
    Learning rate                & $1\times 10^{-3}$           \\
    \hline
    
    \hline
  \end{tabularx}
\end{small}
  \caption{Policy Neural Network Hyperparameters}
  \label{tab:param-nn}
\end{table}

\subsection{Neural Network Policy Training}
We here explain how we obtain policy~$\pi^\text{aux}$ that is used in \S{}\ref{sec:neural-policy} to guide Monte-Carlo Tree Search (MCTS).  We obtain~$\pi^\text{aux}$ optimizing the parameter of a neural network as follows.

Let~$p_\theta(e|s)$ represent a parametric policy, where~$\theta$ is the vector of parameters of a Multi-Layer Perceptron (MLP). We train it as follows. Let us assume we have a training set composed of $z=1,\dots,Z$ exemplary network trajectories. Each example~$\pazocal{G}_z(t)_{t\in[0,T]}$ is a sequence of Time-Expanded Graphs, obtained by sequentially taking actions (i.e., adding time-expaneded edges sequentially). We decompose it into different chunks $\pazocal{G}_z(t)_{t\in[0,T_1]}, \pazocal{G}_z(t)_{t\in[T_1,T_2]},..,\pazocal{G}_z(t)_{t\in[T_x,T]}$. For chunk~$\pazocal{G}_z(t)_{t\in[0,T_1]}$, we extract the last TEG (denote the corresponding state~$s'$) and the second-to-last TEG (denote the corresponding state~$s$). We then recover action~$e$ such that $s\xrightarrow{e} s'$. From all the chunks of all the exemplary trajectories, we collect all such transitions in the form~$s\xrightarrow{e} s'$. Let~$\{s_k \xrightarrow{e_k} s_k^\prime, k=1,\dots,K\}$ be the dataset of all such trajectories.
We embed TEG $\pazocal G(t)$ as a $\pazocal V\times \pazocal V$ matrix, i.e., with as many columns and as many rows as the number of nodes. For simplicity, the embedding of a state~$s(t)=(\pazocal B(t), \pazocal G(t))$ corresponds to the embedding of~$\pazocal G(t)$, irrespective of buffer~$\pazocal B(t)$.\footnote{
Recall that the policy we are constructing here is not the one that fully determines the actions to be applied. The policy we are constructing here is rather guiding the main policy, by favoring graph transitions similar to the transitions observed in the historical dataset. For simplicity, this guidance is just based on graph transitions and not buffer states.
}
In the $(v,v')$-cell, we write the latest departure time among those of the time-expanded edges in~$\pazocal G(t)$. This matrix is the input of the MLP. As output, we have a vector of probabilities, each corresponding to a certain potential action. Let~$p_\theta(e|s)$ denote the output corresponding to the probability of choosing action~$e$. We learn parameters~$\theta$ that maximize the log-likelihood of~$p_\theta(e|s)$ against the observed transitions, i.e.:
\[
\theta^* \in\arg\max_\theta \frac{1}{K}\sum_{k=1}^K \log p_\theta(e_k|s_k)
\]

We set~$p(e|s)=p_\theta(e|s)$.

Table~\ref{tab:param-nn} summarizes the hyperparameters of the policy neural network used in our numerical results. Observe that we could have presented deeper or more complex neural architectures. In our case, however, we verified that this did not lead to improved results. This is because in our transport application, the transport operator would typically have a relatively small dataset of exemplary trajectories, which is not suited to train massive neural architectures.

\subsection{Detailed performance results}
\label{sec:detailed-performance-results}

\begin{figure}
    \centering
\includegraphics[width=.3\linewidth]{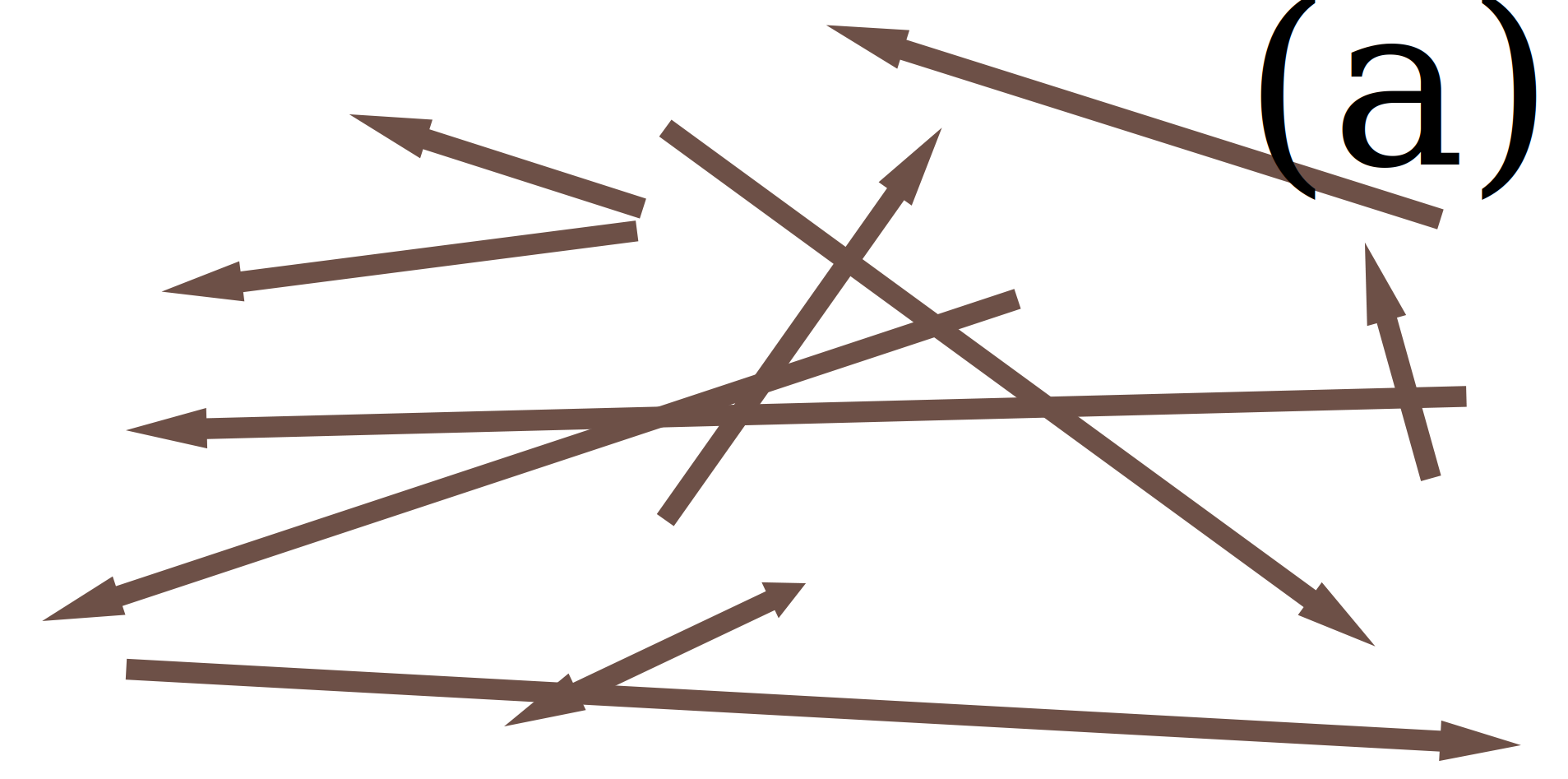}
\includegraphics[width=.3\linewidth]{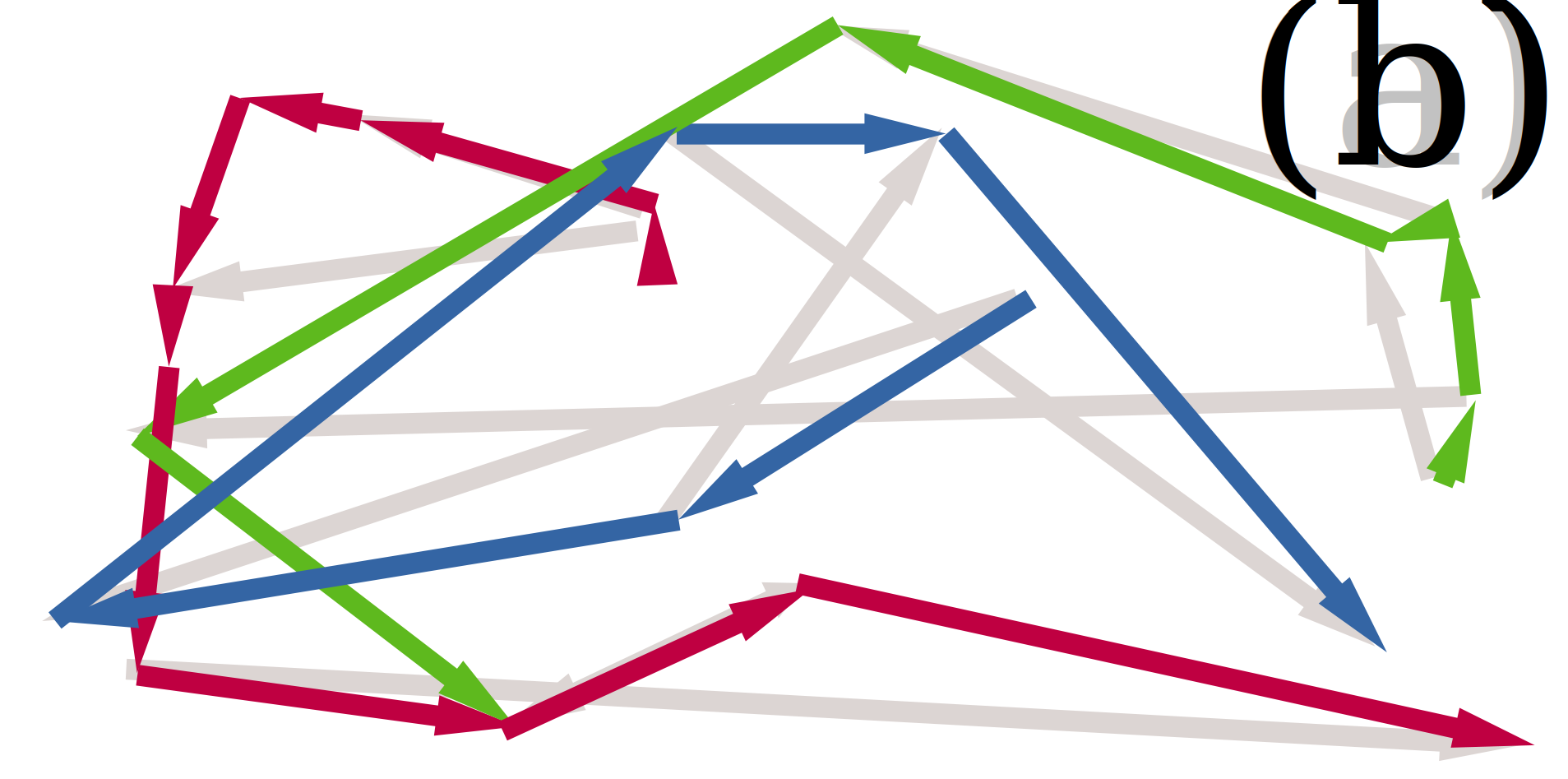}
\includegraphics[width=.3\linewidth]{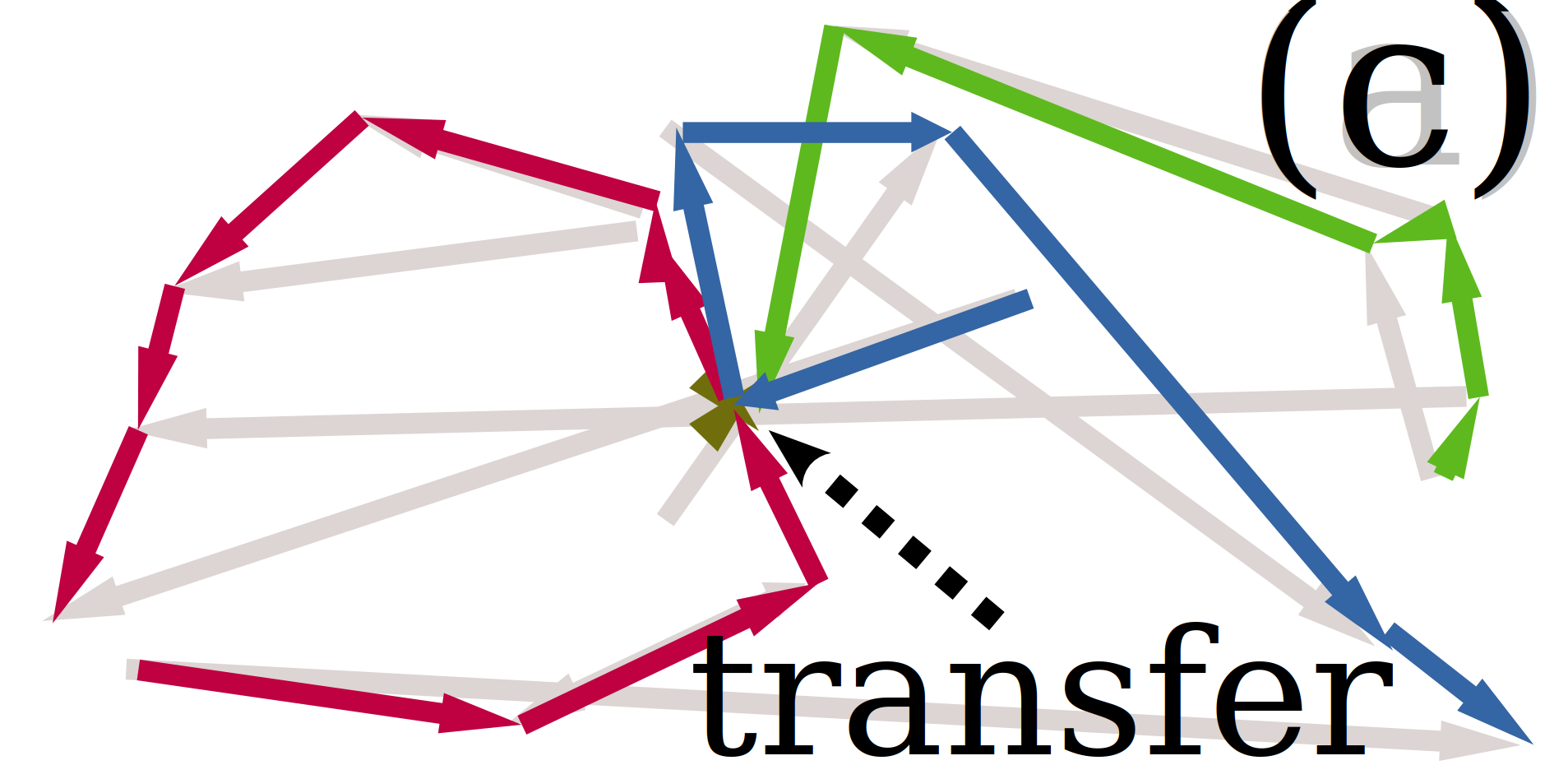}
    \caption{(a)~Trip requests (b)~DVRP-based routes to serve them (c)~Routes issued by our online network design method.}
    \label{fig:example}
\end{figure}

Fig.~\ref{fig:example} conceptually depicts how the network structure that we design naturally induces a synchronism between routes, which enables transfers and reduces miles traveled.

\subsubsection{Impact of the fleet size}

The performance of our method with different fleet sizes is shown in Figure~\ref{fig:3} and Table~\ref{tab:2}. Figure~\ref{fig:3} shows the percentage of unserved requests out of all requests received so far. For any fleet size, the number of unserved requests is quite high in the period starting at 9:00. This is because our system has just started, and we initialize TEG~$\pazocal G(t)$ randomly at~$t=$ 9:00, which means that bus schedule is randomly planned in until time $t+L=9:30$ (Line~\ref{ln:random} of Alg.~\ref{fig:algo_1}). After the simulation runs for a period of time, the percentage of unserved requests out of all requests received so far goes below 9\% with a sufficiently large fleet size. Figure~\ref{fig:3} also illustrates the impact of the proposed neural network policy~$\pi^\text{aux}$ (\S{}\ref{sec:neural-policy}). Its benefit is observed in the initial phase, where it guides Monte-Carlo Tree Search (MCTS) to take directly good actions.
In the rest of the paper, we report and compare the results obtained with the policy-guided MCTS under a fleet size of 40.

It should be noted that, in any dynamic bus system, it is impossible to serve all requests. There can be requests whose origin and destination are very far or in remote places. This kind of trips would likely be unserved also in current real transport systems (and the user would be better-off taking their car). Public transport needs indeed to balance user satisfaction and efficiency, and it cannot thus provide a service tightly tailored to very specific user needs.
As expected, a larger fleet size serves more requests and reduces waiting time (Table~\ref{tab:2}). The table also shows that larger fleet sizes increase average trip time. This might appear surprising, but it is not, for the following reason. Increasing the fleet size enables the system to serve requests that remain unserved under smaller fleets. These additional requests are typically associated with longer O–D distances and in-vehicle times. Therefore, their inclusion mechanically increases the average trip time. A more detailed quantitative analysis of this trade-off will be presented later in Figure~\ref{fig:heat_map_OD_distance_invehicle}.



\begin{figure}[]
  \centering
  \includegraphics[width=0.4\textwidth]{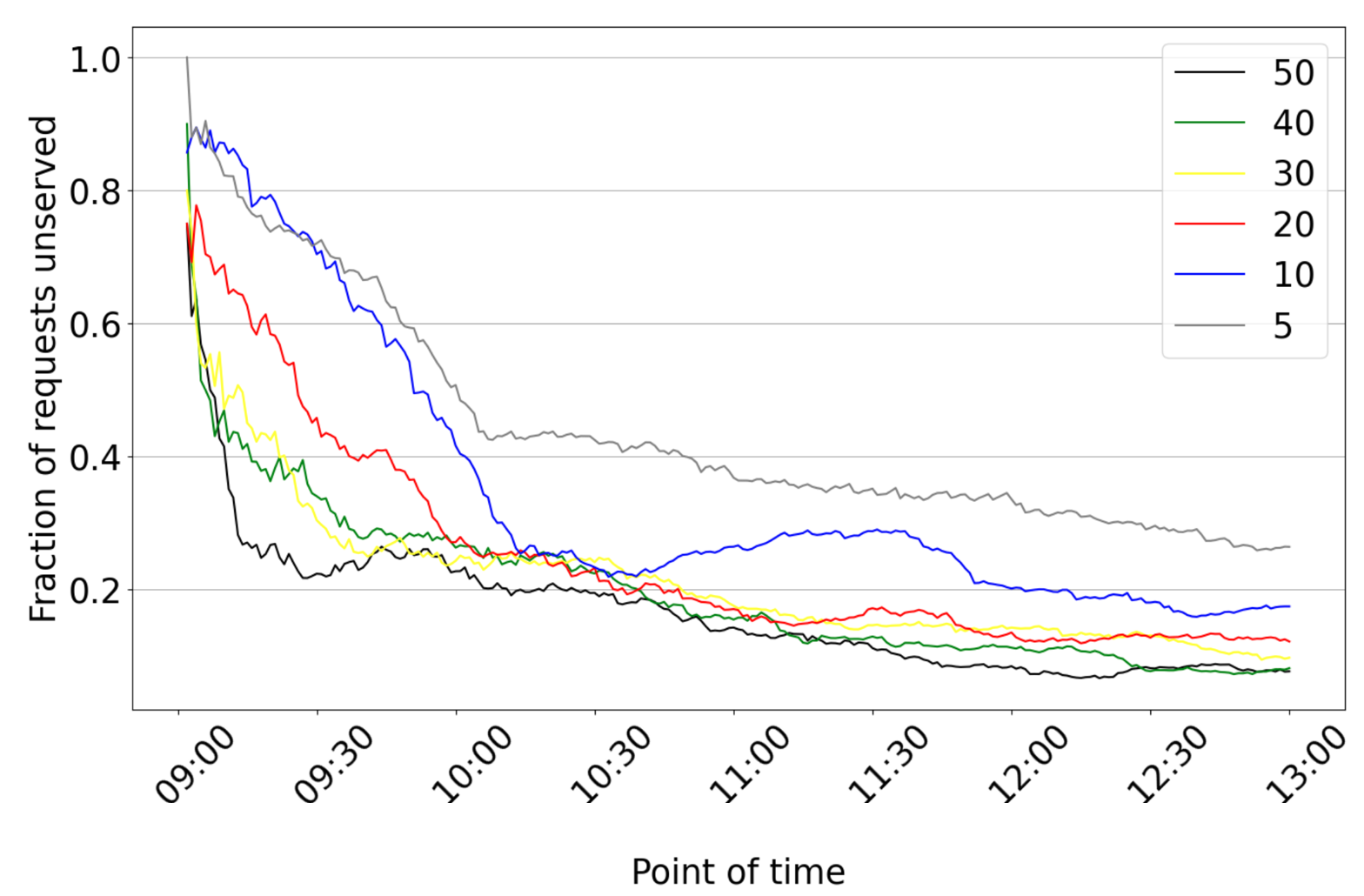}
  \caption{Fraction of requests unserved from 9:00 to 13:00.}
  \label{fig:3}
\end{figure}

\begin{table*}[t]
    \centering
    \begin{tabular}{|c|c|c|c|c|}
        \hline                
                Fleet size 
                 &  Service rate &  Avg. Trip time (mins)&  Avg. Waiting time (mins)\\
         \hline
                  5 &  73.58\%&   27.81 &  16.01 \\ \hline
                  10 & 82.55\%&   27.75 &  15.76 \\ \hline
                  20 & 87.82\%&   29.68 &  15.03  \\ \hline
                  30 & 90.25\%&   29.93 &  14.20  \\ \hline
                  40 & 91.83\% &  31.79 &  14.63 \\ \hline
                  50 & 92.31\% &  32.20 &  13.89   \\ \hline
    \end{tabular}
    \caption{Comparison of results with different fleet sizes }
    \label{tab:2}
\end{table*}

\begin{table*}[t]
    \centering
    \begin{tabular}{|c|c|c|c|c|}
        \hline                
                Fleet size & Avg. Occupancy 
                &  Avg. Stretch w.r.t & Avg. Stretch w.r.t &  Averrage number\\
                & (requests/bus)
                & direct car trip
                & walking trip
                & of transfers
                \\
         \hline
                  5 & 24&1.80& 0.68  & 1.24 \\ \hline
                  10& 14& 1.56& 0.61 &  1.79 \\ \hline
                  20& 9&  1.90&0.72  & 1.85  \\ \hline
                  30& 7&  2.35&0.89 &  1.59 \\ \hline
                  40& 6& 2.62&0.99 &  1.68 \\ \hline
                  50& 5& 2.83&1.07  & 1.50 \\ \hline
    \end{tabular}
    \caption{Performance with different fleet sizes }
    \label{tab:3}
\end{table*}

\begin{figure}[]
  \centering
  \includegraphics[width=1.0\linewidth]{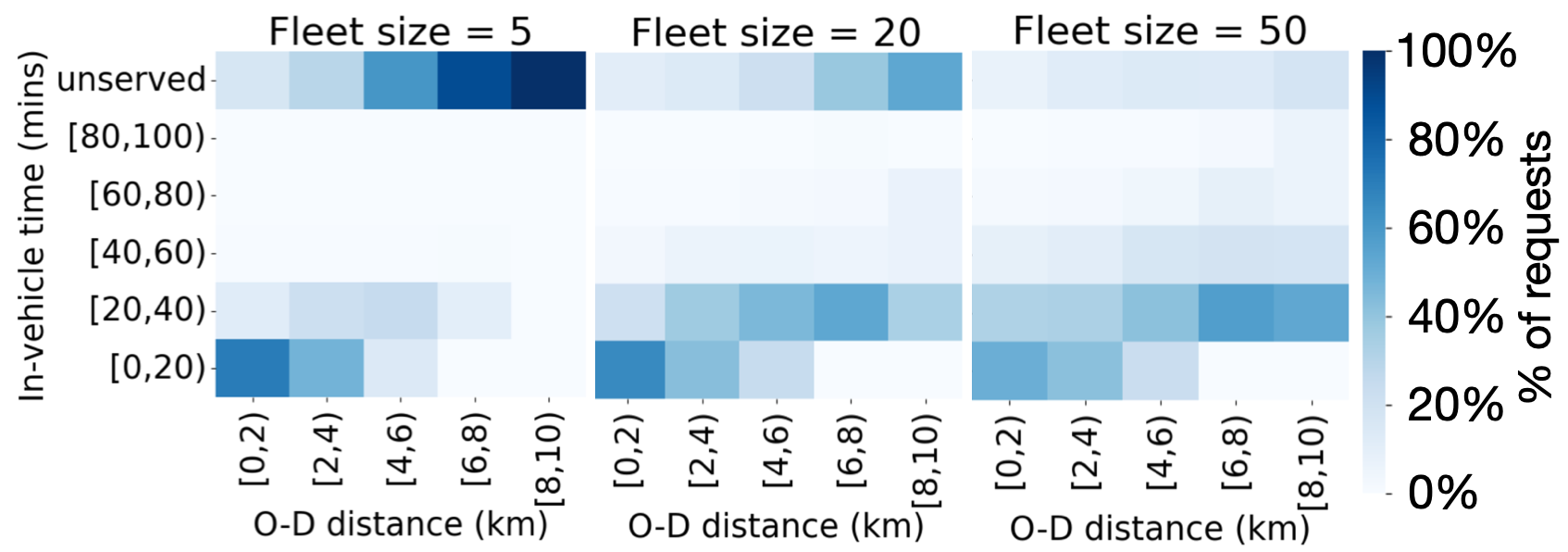}
  \caption{ Fraction of requests served with a certain in-vehicle time over total requests, per each O-D distance interval.}
\label{fig:heat_map_OD_distance_invehicle}
\end{figure}

\subsubsection{Occupancy}
We measure the average occupancy in a bus as follows.

\begin{proposition}
Let $T_\text{in-vehicle}$ denote the average in-vehicle time , $N_\text{served}$ the total number of served request, $]0,T[$ the considered lifespan and~$FS$ the fleet size.
The Average occupancy of a bus is:
$$N_\text{occupancy} = \frac{N_\text{served}}{FS \cdot T} \cdot T_\text{in-vehicle}.$$
\end{proposition}

\begin{proof}
\todo{this proof might be improved}
Factor~$\frac{N_\text{served}}{ T}$ is the rate of requests (req/min) the system picks up. The average rate of requests picked up by one bus is thus~$\frac{N_\text{served}}{FS \cdot T}$.
Requests~$\frac{N_\text{served}}{FS \cdot T}$ picked up at instant $t$ will alight at $t + T_\text{in-vehicle}$ on average, but in the same interval~$]t,t+T_\text{in-vehicle}$ the bus can also pick up number~$\frac{N_\text{served}}{FS \cdot T}\cdot T_\text{in-vehicle}$ of requests. Each bus starts running at time~$0$ and during time interval $]0,0+T_\text{in-vehicle}[$, it picks up on average$\frac{N_\text{served}}{FS \cdot T} \cdot T_\text{in-vehicle}$ requests, and after $t_0 + T_\text{in-vehicle}$, the rate of picks up and drop offs is equal on average, so the average occupancy of the bus  remains the same.   
\end{proof}

Table~\ref{tab:3} shows the average vehicle occupancy in the considered scenario. The stretch of a trip in our system measures the ratio between the in-vehicle time in our system over the time that trip would have taken if it were performed via a direct car trip or via walking, i.e., $\frac{T_{\text{in-vehicle}} \cdot v_{\text{car or walking}}}{O-D\,\text{distance}}$. Table~\ref{tab:3} shows the average across the observed trips.
It is interesting to note that our method adapts to a small fleet, as it is able to increase the level of \emph{sharing}, i.e. the smaller the fleet the more user trips are consolidated (i.e., higher occupancy) and thus the more ``shared'' is each vehicle kilometer traveled.
As fleet sizes increases, the average occupancy decreases, which suggests that the kind of vehicles that should be adopted might change (from large buses to minibuses).

\vspace{0.5cm}
\subsubsection{Impact of the trip distance}
\label{sec:trip-distance}
Figure~\ref{fig:heat_map_OD_distance_invehicle} shows service rates across trips of varying lengths. O–D distances and in-vehicle times are discretized into five intervals. For each O–D distance bin, we compute the fraction of requests served within each in-vehicle time bin (e.g., 0–20, 20–40 minutes). The color scale represents these fractions.
%
For example, the bottom left square color represents 
\begin{small}
$$\frac{\text{served req. with OD dist.}\in [0,2\text{km})\text{ \& trav.time}\in[0,20\text{min}) }{\text{All requests issued with O-D distance}\in\,[0,2\text{km})}.
$$
\end{small}
As expected, larger fleet sizes increase service rates across all trip lengths, leading to fewer unserved requests. Notably, the marginal benefit of increasing the fleet is more pronounced for longer trips.

\begin{figure}[]
  \centering
  \includegraphics[width=1.0\linewidth]{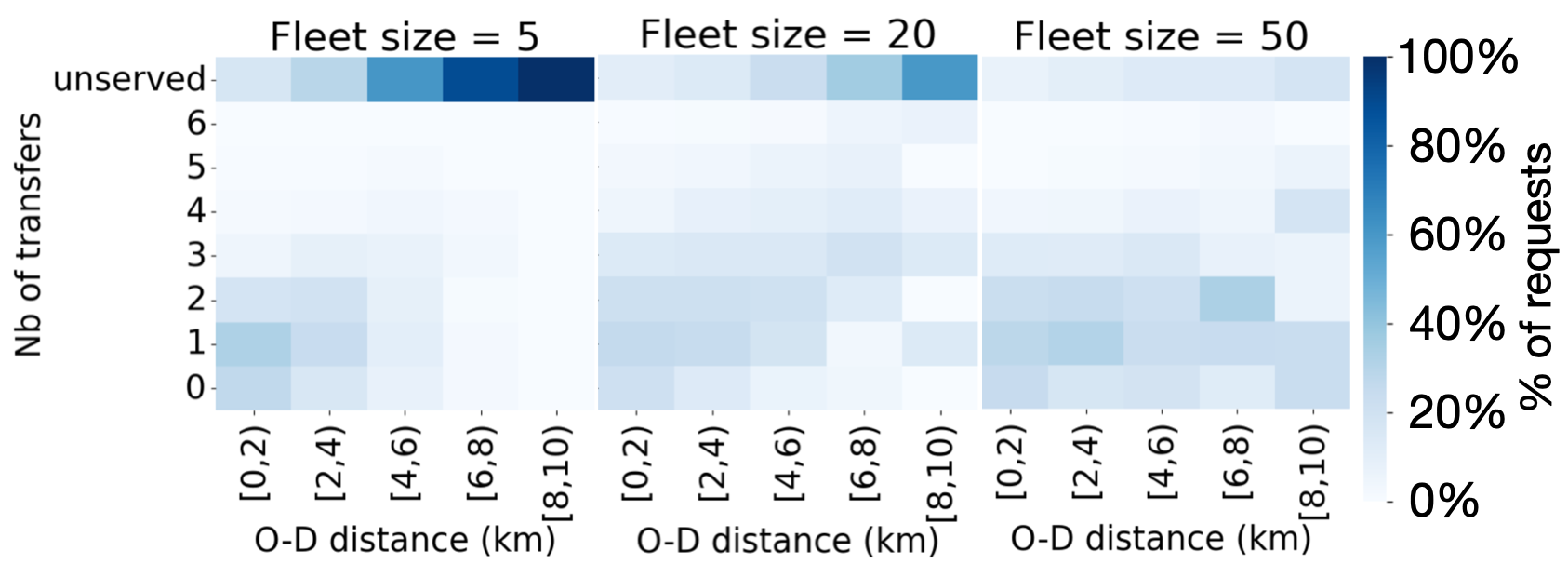}
  \caption{ Fraction of requests experiencing certain number of transfers over total requests, per each O-D distance interval. }
  \label{fig:heat_map_OD_distance_transfer}
\end{figure}

Figure~\ref{fig:heat_map_OD_distance_transfer} represents the number of transfers per trip and is obtained in the same way as Figure~\ref{fig:heat_map_OD_distance_invehicle}. We observe that, with a larger fleet size, more transfers are possible, as the network of bus lines is denser. In a real system, it would be reasonable to limit the number of transfers, which negatively impact user experience. We could easily adapt our MCTS approach to the case in which users accept to do at most 1 or 2 transfers: it would suffice to consider as ``served'' users that can go from origin to destination with no more than those transfers and keep in the ``unserved'' list all the other users. In this way, the MCTS algorithm would collect rewards only for the trips respecting the ``maximum transfers constraint'' and would learn to adjust bus lines accordingly. In our current system, 87.75\% of completed requests are served with at most two transfers and 97.4\% with at most three transfers with 50 buses, indicating that even without an explicit transfer cap the system is already sufficiently convenient for most users.


\section{Application to heavy lift launch vehicles and the $k$-server problem}
\label{sec:other}

\subsection{Complex system of Heavy Lift Launch Vehicles}
\label{sec:heavy-lift}

We now demonstrate how our approach can be used to optimize the sequence of decisions in a large-scale project.
We apply our method to the problem of switching among configurations of a complex system of Heavy Lift Launch Vehicles~(HLLVs)~\cite{DeWeck2007}.
HLLVs are large-scale, modular space transportation systems designed to deliver substantial payloads (typically tens of tons) to orbit, whose architecture involves tightly coupled subsystems (e.g., propulsion, staging, and payload integration) and complex design trade-offs across performance, cost, and reliability~\cite{DeWeck2007}.
In this application, substrate graph~$\pazocal{G}_\text{substr}=(\pazocal{V},\pazocal{E}_\text{substr})$ consists of set~$\pazocal{V}$ of potential configurations and~$\pazocal{E}_\text{substr}\subseteq\pazocal{V}\times\pazocal{V}$ of edges. An edge (which also corresponds to a potential action) represents the possibility of switching between two configurations. When we decide on an action, which is to switch from one configuration to another, environment~$\textit{Env}$ generates a certain cost suffered by the system (opposite of reward~$r$), which depends on its current configuration. Contest~$cx(t)$ is the accumulated cost up to~$t$. Having defined states, actions, and reward, we can use our method to design the evolution of configuration switches of the complex system, in real time, with the aim to minimize accumulated stochastic cost.

The authors of~\cite{DeWeck2007} used a Time-Expanded Graph (TEG) to store the cost of all state transitions of a complex system and chose the optimal state-path from an exhaustive enumeration of all possible paths. Eq.(7) of~\cite{DeWeck2007} shows that $ \#\textit{paths}=C^T$, where $C$ is the number of states (i.e., configuration), and $T$ is the time horizon expressed as the number of time slots. They are therefore constrained to small problem sizes: $C=4$, $T=4$, and $C^T =256$ paths. 

We instead optimize the sequence of configurations as a network, using our online network design approach. We do not need any buffer, so the state~$s(t)$ is simply the sequence of visited configurations up to time~$t$. Intervention times, in which a switch from one configuration to another can occur, correspond to the ``stages'' of~\cite{DeWeck2007}. Switching from~$s(t_n)$ to~$s(t_{n+1})$ means to change the configuration from the last configuration of~$s(t_n)$, which we call~$v$ to another configuration~$v'$. This action is represented by time-expanded edge~$e_n=(t_n, v,v')$. When the action is taken, environment~$\textit{Env}'$ generates a configuration-switching cost~$c_{v,v',\omega}$. The reward is~$r_\omega(t_n, e_n | s(t_n))=-c_{v,v',\omega}$, as in~\eqref{eq:reward-cost}. The goal is to minimize the expected value of the cumulative cost across the entire duration of the project. For compliance with~\cite{DeWeck2007}, we do not add any decision lag, i.e., we impose~$L=0$. This does not contradict Prop.~\ref{prop:timing}, because the goal is to minimize cumulative cost rather than maximizing requests served.

Our network-based approach allows us to run the same problem of~\cite{DeWeck2007}, with a much larger state space, i.e., $C=20$, $T=20$. In this case, they would need to evaluate $\#\textit{paths}= 20^{20}$ ($ \approx 10^{26}$) paths, which is infeasible. Our method can instead find satisfying solutions in just about 16 seconds. We test 20 distinct realizations~$\omega\in\Omega$, obtaining an average improvement of 29\% with respect to the baseline (which is the best static system state, which can be found by comparing all~$C$ configurations).

\subsection{The $k$-server problem}
\label{sec:k-server}
The $k$-server problem is a well-known online optimization problem. $k$ servers are located on a graph, requests arrive one by one at some nodes~$v\in\pazocal V$. For each request, one server should go to the request's location to serve it~\cite{koutsoupias2009k}. An action at time~$t_n$ corresponds to moving a server~$l\in\pazocal L=\{1,\dots,k\}$  from a node~$v\in\pazocal V$ to another node~$v'\in\pazocal V$, which generates negative reward~$r_\omega(t_n, e_n | s(t_n))=-c_{v,v',\omega}$, where~$c_{v,v',\omega}$ is the cost incurred in the movement. Such a cost might represent, for instance, the monetary cost incurred to perform such a movement (e.g., energy spent, payment of an employee).
The goal is to minimize the expected value of the cumulative cost generated by the movement of all the servers. Every incoming request needs to be processed immediately. Therefore, for the same reason as in \S{}\ref{sec:heavy-lift}, we can set~$L=0$. Algorithms for the $k$-server problem, such as greedy algorithms, have been analyzed in the online algorithm literature in an adversarial setting, i.e., under worst-case scenarios: upper bounds on the ratio between the cost resulting from such algorithms and ideal optimal baselines are calculated, under the most unfavorable request sequence. However, such algorithms may arbitrarily perform poorly in terms of expected cost.

In contrast, since we aim to minimize the expected cost instead, we do not need to consider the worst-case request sequence, and we can make decisions based on commonly observed request patterns. We thus train a prediction model~$\textit{Env}'$ (see \S{}\ref{sec:prediction} on previously observed sequences of requests to simulate future requests, which will be used to determine the best actions, within Monte Carlo Tree Search~(\S{}\ref{sec:MCTS}). We compare our approach with the commonly used greedy algorithm for the $k$-server problem, which refers to moving the nearest server to the request location each time. In the experiment, we consider a square area from (0,0) to (100,100), initialize $k$ = 3 servers, and generate 100 requests following a spatial-temporal Poisson process (most requests appear in a central region, while others are scattered uniformly across the map.
The time between requests is drawn from an exponential distribution with $\lambda = 1$). The results show that our algorithm has an average of 14.18\% less distance traveled by the servers, compared to the greedy algorithm (average value across $|\Omega|=$50 realizations).

\end{appendices}
\fi 

\end{document}